%% file: main.tex
\documentclass[10pt,twocolumn,letterpaper]{article}

\usepackage{cvpr}              

\input{preamble}

\definecolor{cvprblue}{rgb}{0.21,0.49,0.74}
\usepackage[pagebackref,breaklinks,colorlinks,allcolors=cvprblue]{hyperref}

\title{Learning 4D Panoptic Scene Graph Generation from Rich 2D Visual Scene}

\author{
Shengqiong Wu$^1$, \quad Hao Fei$^1$\thanks{Corresponding author.}, \quad Jingkang Yang$^2$, \quad Xiangtai Li$^{2}$ \\ 
Juncheng Li$^3$, \quad Hanwang Zhang$^2$, \quad Tat-seng Chua$^1$ \\
$^1$National University of Singapore \quad $^2$Nanyang Technological University
\quad $^3$Zhejiang University \\
{\tt\small swu@u.nus.edu}, \quad {\tt\small \{haofei37, dcscts\}@nus.edu.sg}, \quad {\tt\small xiangtai94@gmail.com}\\
{\tt\small \{jingkang001, hanwangzhang\}@ntu.edu.sg}, \quad {\tt\small junchengli@zju.edu.cn}
}

\begin{document}
\maketitle

\begin{abstract}
The latest emerged 4D Panoptic Scene Graph (4D-PSG) provides an advanced-ever representation for comprehensively modeling the dynamic 4D visual real world.
Unfortunately, current pioneering 4D-PSG research can primarily suffer from data scarcity issues severely, as well as the resulting out-of-vocabulary problems; also, the pipeline nature of the benchmark generation method can lead to suboptimal performance.
To address these challenges, this paper investigates a novel framework for 4D-PSG generation that leverages rich 2D visual scene annotations to enhance 4D scene learning.
First, we introduce a 4D Large Language Model (4D-LLM) integrated with a 3D mask decoder for end-to-end generation of 4D-PSG.
A chained SG inference mechanism is further designed to exploit LLMs' open-vocabulary capabilities to infer accurate and comprehensive object and relation labels iteratively.
Most importantly, we propose a 2D-to-4D visual scene transfer learning framework, where a spatial-temporal scene transcending strategy effectively transfers dimension-invariant features from abundant 2D SG annotations to 4D scenes, effectively compensating for data scarcity in 4D-PSG.
Extensive experiments on the benchmark data demonstrate that we strikingly outperform baseline models by a large margin, highlighting the effectiveness of our method.
The project page is \url{https://sqwu.top/PSG-4D-LLM/}.
\end{abstract}

\vspace{-4mm}

\section{Introduction}
\label{sec:intro}

\begin{figure}[!t] 
\centering
\begin{subfigure}{\linewidth}
    \centering
    \includegraphics[width=0.9\linewidth]{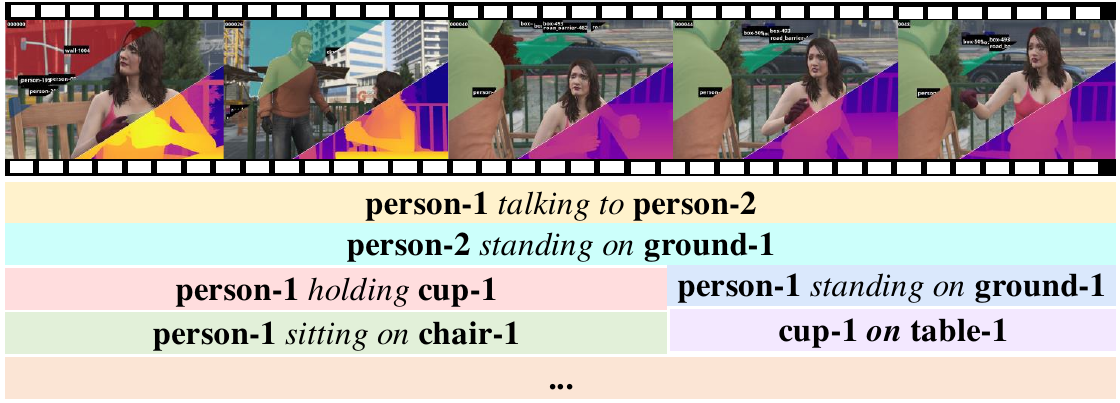}
    \caption{4D panoptic scene graph (4D-PSG) generation. Each frame of scenes has aligned RGB and depth images with panoptic segmentation annotations. The SG is represented as triplets shown bellow.
    }
    \label{fig:intro-a}
\end{subfigure}
\vspace{2mm}
\begin{subfigure}{\linewidth}
    \centering
    \fontsize{8}{8.5}\selectfont
    \setlength{\tabcolsep}{1.1mm}
    \begin{tabular}{llccccc}
        \hline
       \textbf{Modality} &  \textbf{Dataset} & \textbf{\#Obj.} & \textbf{\#Rel.} & \textbf{Triplets} & \textbf{Instances}\\ 
        \hline
    \rowcolor{lightlightgrey} 2D Image & VG~\cite{VG-KrishnaZGJHKCKL17}	&	5,996 &	1,014 &	1,683,231 &	108,077 \\
    Video & AG~\cite{AG-JiK0N20}	&	36 &	25 &	772,013 &	288,782  \\
    \rowcolor{lightlightgrey} 3D & 3DSG~\cite{3DDSG-WaldDNT20} & 528 & 39 & 546,956 & 1,335 \\
    \multirow{2}{*}{4D} & PSG4D-GTA~\cite{4d-psg-yang20244d} & 35  & 43 & 728  & 67 \\ 
     & PSG4D-HOI~\cite{4d-psg-yang20244d} & 42 &  15 & 29,375  & 2,973 \\
    \hline
    \end{tabular}
    \caption{The statistics of SG dataset in image, video, 3D, and 4D modalities. 
    ``\#Obj.'' and `\#Rel.' denote the object and relation number, respectively.}
    \label{fig:intro-b}
\end{subfigure}
\vspace{1mm}
\begin{subfigure}{\linewidth}
    \centering
    \includegraphics[width=1\linewidth]{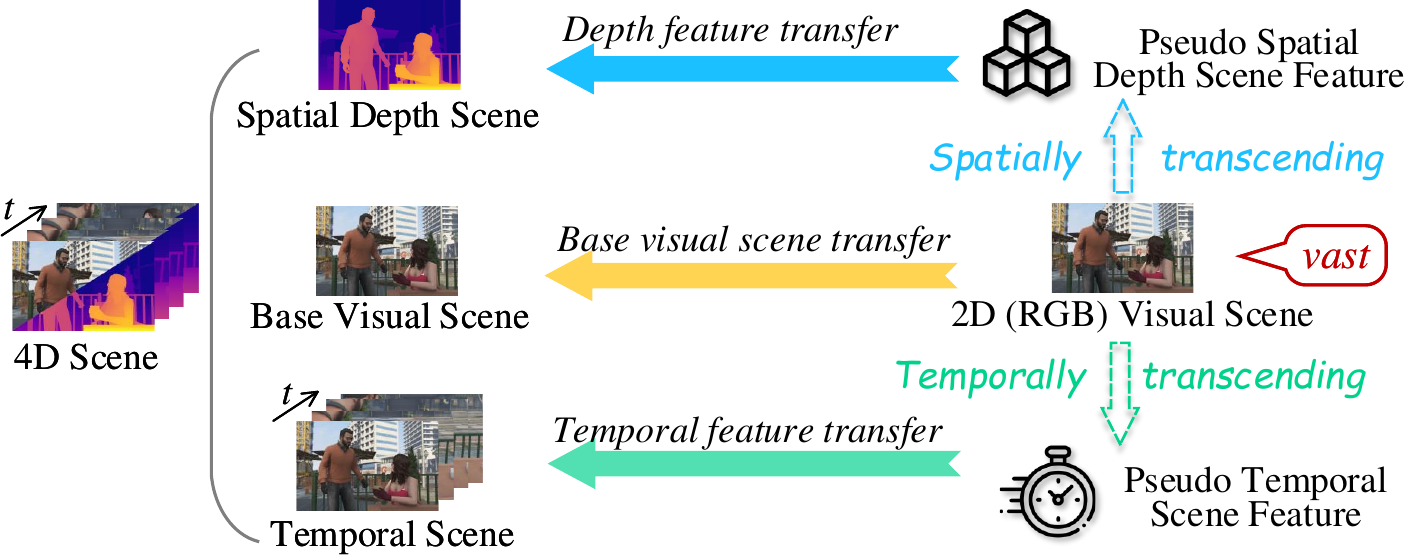}
    \caption{We propose transferring the visual scene features from existing rich 2D data to the 4D-PSG task.
    }
    \label{fig:intro-c}
\end{subfigure}
\vspace{-7mm}
\caption{(a) Illustration of 4D-PSG, (b) SG dataset statistics, and (c) motivation for 2D scene transfer learning.}
\label{fig:image_table}
\vspace{-6mm}
\end{figure}

\vspace{-2mm}
Achieving a comprehensive semantic understanding of real-world scenes is crucial for developing AI systems such as autonomous driving and robotics.
Scene graphs (SGs) \cite{JohnsonKSLSBL15,VG-KrishnaZGJHKCKL17} have thus been proposed as a means to represent scene semantics, with objects and their semantic relations neatly described in a highly structured format.
SG representations exhibit superior capabilities in depicting scenes and play a significant role in numerous downstream visual tasks \cite{GQA-HudsonM19,ShiZL19,abs-2211-11138}. 
Correspondingly, SG generation has garnered extensive research attention over the past decades \cite{SGTR-LiZ022,Pair-Net-10634834,VidVRD-shang2017video,VL-SAT-WangCZ0TS23}.
Initially, research \cite{Motifs-ZellersYTC18,VCTree-TangZWLL19,RelTR-CongYR23,EGTR-ImNPLP24} was confined to SG for static images, aiming to depict language-based scenes within images.
Subsequently, it was extended to video scenes to facilitate dynamic temporal visual scene modeling \cite{PVSG-YangPLGCL0ZZLL23,OED-WangLCL24}.
Meanwhile, Panoptic Scene Graphs (PSGs) \cite{PSG-YangAGZZ022} expanded the definition to include pixel-level object localization and grounding, providing a more fine-grained and comprehensive scene understanding.
To further overcome the dimensional limitations of 2D image visuals, 3D-SGs \cite{3DDSG-WaldDNT20} were introduced, wherein visual objects are three-dimensional and engage in spatial interactions.
Given that the real world is a dynamically 3D spatiotemporal environment, existing SG definitions might fail to meet real-world requirements. 
Thus, 4D Panoptic Scene Graphs (4D-PSGs) \cite{4d-psg-yang20244d} have recently been proposed, which further enhance the definition of 3D-SGs by incorporating additional temporal relationships, as illustrated in Fig. \ref{fig:intro-a}, achieving dynamic high-level visual modeling in four dimensions.

While Yang et al. (2023) \cite{4d-psg-yang20244d} pioneered 4D-PSG with a benchmark solution, we identify several non-trivial challenges in their research that may potentially hinder the further development of this task.
\textbf{First}, the primary issue lies in the scarcity of the 4D-PSG benchmark dataset introduced in their work, as shown in Fig. \ref{fig:intro-b} of comparing current SG data across 2D, Video, 3D, and 4D, where 4D-PSG has only 1.7\% 2D-SG data.
Worse still, part of the data is either synthetic, or suffers from domain bias (predominantly limited to indoor scenes), lacking comprehensive instances of real-world scenarios. 
Such data sparsity severely impedes the learning of a satisfactory generation model.
\textbf{Second}, similarly constrained by data scarcity, the out-of-vocabulary problem \cite{KochVCHR24,chen2024expanding,Wu0X24} is significantly exacerbated in 4D-PSG generation. 
Yet the SG generation method proposed in \cite{4d-psg-yang20244d} fails to account for open-vocabulary settings, supporting only a predefined set of approximately 50 object classes and 40 relation classes. 
This limitation substantially restricts the types of scenes the model can handle in real-world environments, limiting its practical application value.
\textbf{Last but not least}, regarding the 4D-PSG generation method, \cite{4d-psg-yang20244d} proposes a pipeline system, i.e., from 4D panoptic segmentation to relation determination. 
Unfortunately, such a cascade architecture inevitably introduces errors, diminishing the overall performance.

This work is dedicated to addressing all the aforementioned challenges.
On the one hand, we propose a novel 4D-PSG generator.
We build upon a 4D Large Language Model (4D-LLM) and jointly integrate a 3D mask decoder, enabling the system to comprehend the given 4D scene input and end-to-end output panoptic 3D objects with all corresponding spatial-temporal relations, i.e., 4D-PSG.
By leveraging an LLM as the backbone, our system offers a much more powerful scene recognition capacity, delivering superior task performance compared to the traditional specialist models.
Meanwhile, we design a \emph{chained SG inference} mechanism, which fully leverages the LLM's robust open-vocabulary generalizability to iteratively infer SG object and relation labels that provide the best accurate and reasonable descriptions.
On the other hand, to mitigate the data scarcity, we introduce \textit{learning dimension-invariant visual scene features from the vast and rich 2D SG annotation data as compensatory supervision, and transfer them to 4D scenes for 4D-PSG learning}, as depicted in Fig. \ref{fig:intro-c}.
Technically, we introduce a \emph{2D-to-4D visual scene transfer} (in short, D\textsuperscript{2$\to$4}-VST) learning framework, which is based on a \emph{spatial-temporal 2D-to-4D scene transcending} mechanism, learning the dimension transcending modes of 2D-to-3D and 3D-to-4D respectively using a limited amount of 4D data.
Afterward, this dimension transcending is unsupervisedly performed on a massive corpus of 2D image annotations of SGs, obtaining rich 4D visual features.

We conduct extensive experiments on the 4D-PSG benchmark data, and the results demonstrate that our system strikingly outperforms the baselines by a large margin across all metrics and scenarios.
Further analysis reveals the effectiveness of our proposed D\textsuperscript{2$\to$4}-VST learning mechanism, and the LLM-based chained SG inference mechanism significantly enhances open-vocabulary capabilities.
Overall, our contributions are as follows:
\circlenum{I} proposing a 4D-LLM based model for end-to-end 4D-PSG generation;
\circlenum{II} designing a novel chained SG inference mechanism to address the out-of-vocabulary problem effectively;
\circlenum{III} developing a 2D-to-4D visual scene transfer learning framework to obtain 4D visual scene supervisions from tremendous amount of 2D resource, solving the data scarcity issue;
\circlenum{IV} empirically improving 4D-PSG with a large margin and significantly advancing the research in this direction.

\vspace{-2mm}
\section{Related Work}
\vspace{-1mm}

SGs \cite{JohnsonKSLSBL15,DSGG-Hayder024} have attracted extensive research attention due to their concise definitions and capabilities in representing scenes, achieving applications across various practical tasks \cite{,fei-etal-2023-scene,fei2024dysen,WangMZYF24,0001W0ZZLH24}.
Historically, SGs have been developed across various modalities, including images \cite{VG-KrishnaZGJHKCKL17,PSG-YangAGZZ022,wu-etal-2023-cross2stra,OMGSeg,li2023transformer}, videos \cite{VidOR-shang2019annotating,VG-KrishnaZGJHKCKL17,Zhao-mm-23,VidVRD-shang2017video,PVSG-YangPLGCL0ZZLL23}, 3D environments \cite{3DDSG-WaldDNT20, CCL-3DDSG-ChenWLLWH24,VL-SAT-WangCZ0TS23}, and 4D data \cite{4d-psg-yang20244d}, even textual data \cite{SchusterKCFM15,FACTUAL-LiCZQHLJT23}.  
Subsequent research has extended SG representations to diverse scene settings, such as panoptic SGs \cite{PSG-YangAGZZ022,PVSG-YangPLGCL0ZZLL23}, ego-view SGs \cite{ZhangYHK0C22,RodinF0TF24}, etc.  
While numerous SG generation approaches have also been developed, they generally require two key predictions: (a) scene object detection and (b) semantic relation recognition. 
This work serves as a follow-up to the 4D-PSG task, aiming to enhance it in several essential aspects.

Our work is closely related to research on weakly-supervised SG generation \cite{KimYJIM0P24,abs-2404-02527,0005ZX0X21}.  
In many cases, annotating SGs is costly and can introduce domain bias—for example, most SGs \cite{3DDSG-WaldDNT20} are often limited to indoor scenes.  
Therefore, weakly supervised learning provides significant assistance in addressing data scarcity.  
Previous related work primarily focuses on utilizing abundant image-text pairs without SG annotations to obtain additional supervisory labels.  
For instance, \cite{LouHLZ22,YeK21} parses the linguistic structure of textual captions corresponding to images as pseudo-SG labels.
More recently, \cite{KimYJIM0P24} employs LLMs to generate captions and corresponding syntactic parses for images and then generates pseudo-SG labels through phrase grounding.  
Annotating 4D SGs is inherently constrained by the availability of datasets, leading to the data scarcity problem in 4D-PSG.
To address this, we consider a novel perspective: learning rich, modality-consistent visual scene features from the vast and abundant 2D SG annotation data as supplementary signals, and transferring them to 4D scenes for 4D-PSG.

Besides, we attempt to solve another significant challenge in SG generation, the out-of-vocabulary problem \cite{CCL-3DDSG-ChenWLLWH24,chen2024expanding,Wu0X24}, where models can only assign predefined labels for objects and relations, which hampers broader applicability in diverse real-world applications. 
To this end, previous methods \cite{LiZLC024,0056PY0MC23,Wu0X24,ChenWLLWH24} leverage aligned vision-language models, such as CLIP \cite{CLIP-RadfordKHRGASAM21}, to learn object and relation alignments between image and language data, thereby enabling the handling of unseen objects and relations.  
In contrast, we leverage recent LLMs, fully utilizing their powerful open-vocabulary generalizability \cite{llama2-abs-2307-09288,YangZXLHL24,ZhangL0YWJLY24} by prompting LLMs to iteratively infer SG object and relation labels that best provide accurate and reasonable open-vocabulary descriptions.

\begin{figure*}[!th]
    \centering
    \includegraphics[width=0.98\textwidth]{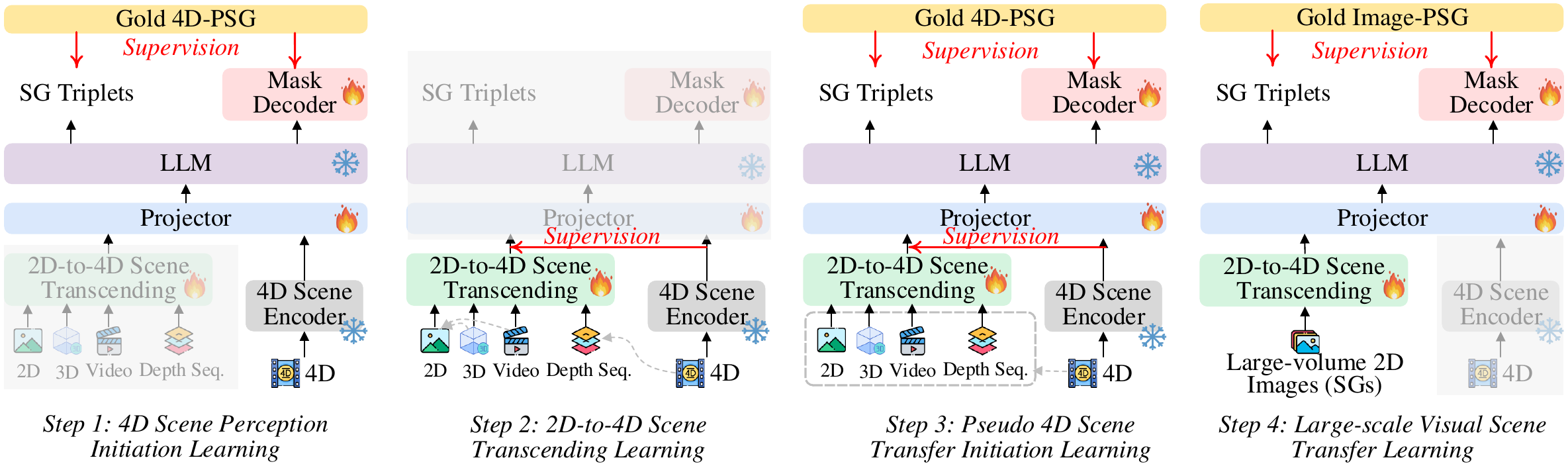}
    \vspace{-2mm}
    \caption{Overview of 2D-to-4D visual scene transfer learning mechanisms for 4D-PSG generation, including 4 key steps.
    }
    \label{fig:frame}
\vspace{-4mm}
\end{figure*}

\vspace{-1mm}
\section{Scene Graph Parsing with 4D-LLM}
\vspace{-1mm}

\paragraph{Problem Formulation.}
Given a 4D scene, specifically represented as a sequence of RGB-D frames $\mathcal{I} \in \mathbb{R}^{T \times H \times W \times 4}$, where $T$ denotes the number of frames and each frame has dimensions $H \times W \times 4$, our objective is to generate a dynamic panoptic SG $\mathcal{G} = \{\mathcal{O}, \mathcal{M}, \mathcal{R}\}$. 
Here, $\mathcal{O} = \{\bm{o}_n\}_{n=1}^{N}$ are the set of objects present in the scene, and $\mathcal{M} = \{\bm{m}_n\}_{n=1}^{N}$ denotes the corresponding binary mask tubes, where $\bm{m}_i \in \{0,1\}^{T \times H \times W}$ tracks the spatial extent of object $o_i$ over time $T$.
The relation set $\mathcal{R}=\{\bm{r}_k\}_{k=1}^{K}$ defines interactions between objects, with each $\bm{r}_k$ linking a subject and an object through a predicate class over a specific time period $(t_s, t_e)$.

\subsection{Model Architecture}
\vspace{-1mm}
The framework is shown in Fig. \ref{fig:frame} (e.g., subfigure step 1).
A 4D scene encoder is applied to perceive the 4D scene input, followed by a projector to bridge LLM and the encoder, enabling LLM to comprehend and reason the 4D scene.
In addition to the SG triplets for the LLM to generate, a mask decoder is used to produce mask tubes that track objects over time. 
To further enhance 4D scene understanding, we implement 2D-to-4D visual scene transfer learning with a spatial-temporal 2D-to-4D scene transcending mechanism.

Specifically, given dual inputs of RGB and depth images of the 4D scene, we use ImageBind \cite{imagebind-GirdharELSAJM23} as the 4D scene encoder for each modality separately, followed by an aggregator, inspired by \cite{ChenLWWQLZ20}, to efficiently fuse feature from all modalities.
Next, since our scene understanding focuses primarily on object-level and relation-level comprehension and we seek to optimize inference efficiency, we follow the approach in \cite{Chat-UniVi-0001TZC024} to merge the resulting representations spatially and temporally. 
The merged features are then passed through an MLP projector layer, transforming the embeddings into the language space for LLM comprehension. 
We instantiate the LLM with LLaMA2 \cite{llama2-abs-2307-09288}, and leverage the LLM to output textual relation triplets sequences for SG generation.
Also, we introduce a signal token ``\texttt{[Obj]}'' to trigger object segmentation. 
Therefore, the output sequence takes the form: ``$o_i$ \texttt{[Obj]} $r_k$ $o_j$ \texttt{[Obj]} $t_s$ $t_e$''.
At the backend, we employ SAM2 \cite{sam2-abs-2408-00714} as a 3D mask decoder, which takes both the hidden states of the ``\texttt{Obj}'' tokens and the original RGB image frames as input. 
To ensure compatibility with SAM2, a linear projector is applied to first project the hidden states to match the dimensions of SAM2's prompt embedding. 
The projected hidden states are then used as prompt embeddings for SAM2.

\vspace{-2mm}
\subsection{Chained Scene Graph Inference}
\vspace{-1mm}

Instead of directly outputting pure SG triplets, we leverage the generalization capabilities of the LLM and its in-context reasoning \cite{tefnik-kadlcik-2023-context,abs-2311-16492} ability to address out-of-vocabulary issues in SG generation. 
Inspired by recent advances in the Chain-of-Thought mechanism \cite{Wei0SBIXCLZ22}, we design a chained SG inference approach that guides the LLM to refine the most suitable object and relation labels step-by-step iteratively.
Emulating human reasoning, which involves first analyzing an object's characteristics based on prior knowledge, then classifying the objects, and finally identifying and specifying relationships, we break down the SG inference into four stages: 
\textbf{First}, we prompt it to provide a descriptive definition of each object before assigning it a category.
\textbf{Second}, based on the identified objects, the LLM analyzes which objects may have semantic relations with one another.
\textbf{Third}, for potential object pair, the LLM describes the relationship with the most precise language possible.
\textbf{Finally}, for each specified relation, the LLM determines its duration or time span.
We show the detailed prompt below.

\vspace{-2mm}
\begin{tcolorbox}[breakable, fontupper=\customfont, title=Chained Scene Graph Inference]
\vspace{-2mm}
{\small
\textbf{Input Data}: 4D Scene, the duration \\
\textbf{Instruction}: 
You are a scene expert with professional skills in generating an SG triplets sequence. You follow these four detailed steps to ensure a logical, step-by-step approach to SG generation: \\

\textbf{\color{ired}{Inference stage 1: Object Description and Categorization}}. 
For each object in the scene, do not immediately output its name. Instead, start by describing each object in detail.
Provide a description of each object based on its appearance, shape, structure, and any unique characteristics observed in the scene. 
After giving a detailed description, assign a category to the object that best fits the objects (e.g., ``person'', ``table'', ``chair'', etc.).\\
\textbf{\color{ired}{Expected Output}}: (description, object$_1$), $\cdots$ \\

\textbf{\color{blue}{Inference stage 2: Semantic Relation Identification}}. 
Based on the identified objects, analyze which pairs of objects may have semantic relations. Consider spatial positioning, interactions, and any logical connections that might exist between them.
Identify only pairs that have a meaningful relationship and briefly explain why these pairs might be related.\\
\textbf{\color{blue}{Expected Output}}: (object$_i$, object$_j$), $\cdots$ \\

\textbf{\color{igreen}{Inference stage 3: Precise Relation Description}}
For each object pair identified in Step 2, describe the exact nature of the relation between the two objects as precisely as possible.
Use clear, concise language to specify the relation type (e.g., "sitting on," "holding," "near," etc.) and provide additional context if necessary to ensure the relation is unambiguous.\\
\textbf{\color{igreen}{Expected Output}}: (object$_i$, relation$_k$ object$_j$), $\cdots$ \\

\textbf{\color{ipurple}{Inference stage 4: Temporal Span Determination}}
For each identified relation, determine its duration or time span. Indicate if the relation is continuous, occurs intermittently, or exists only at a specific moment within the scene.
Use a numerical value for the duration, such as a time interval (e.g., (0.1, 0.7) ) \\
\textbf{\color{ipurple}{Expected Output}}: (object$_i$, relation$_k$ object$_j$, start\_time, end\_time), $\cdots$ \\

\textbf{Final Output Format}:
For each object pair and relation, generate SG triplets in the following format:\\
\textbf{\color{black}{Expected Output}}: (object$_i$, relation$_k$ object$_j$, start\_time, end\_time), $\cdots$
}
\end{tcolorbox}

\vspace{-3mm}
\section{2D-to-4D Visual Scene Transfer Learning}
\vspace{-2mm}

\begin{figure}[!t] 
\centering
\begin{subfigure}{\linewidth}
    \centering
    \includegraphics[width=1.0\linewidth]{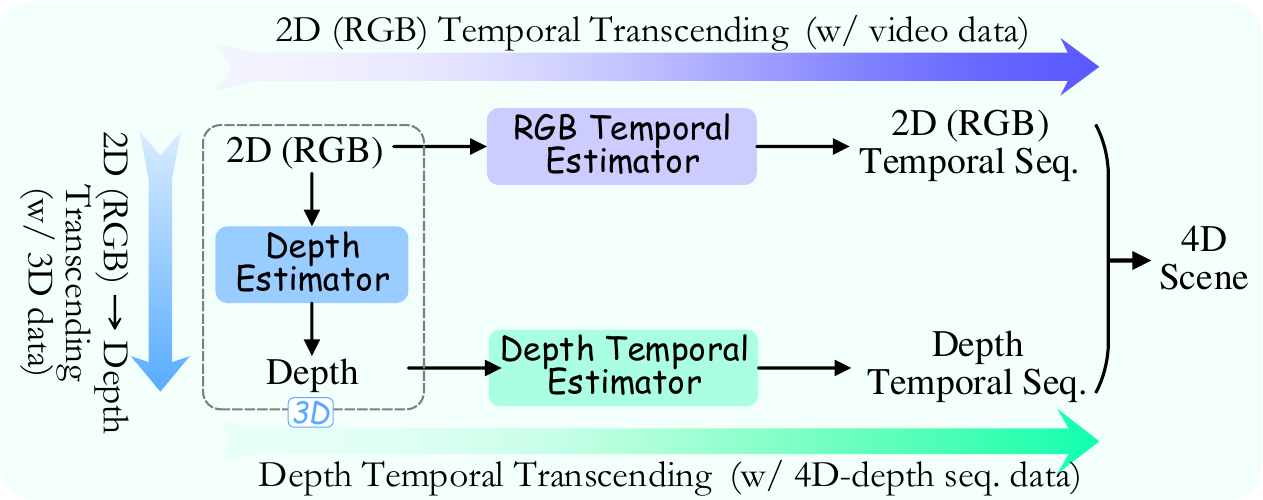}
    \caption{The spatial-temporal 2D-to-4D transcending mechanism.}
   \label{fig:2d-to-4d-overall}
\end{subfigure}
\vspace{2mm}
\begin{subfigure}{\linewidth}
    \centering
    \includegraphics[width=1.0\linewidth]{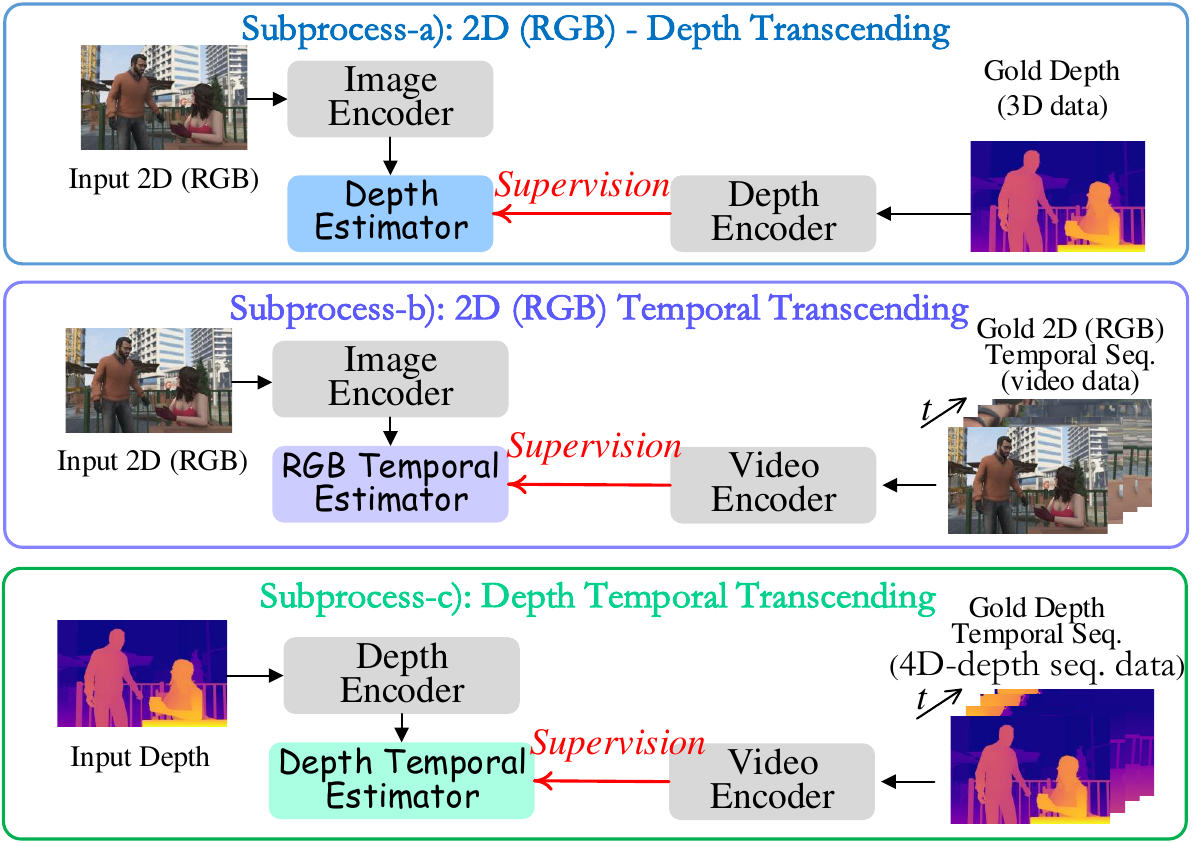}
    \caption{The training stage of scene transcending (refer to step 2\&3 in Fig. \ref{fig:frame}).}
   \label{fig:2d-to-4d-detail}
\end{subfigure}
\vspace{2mm}
\begin{subfigure}{\linewidth}
    \centering
    \includegraphics[width=1.0\linewidth]{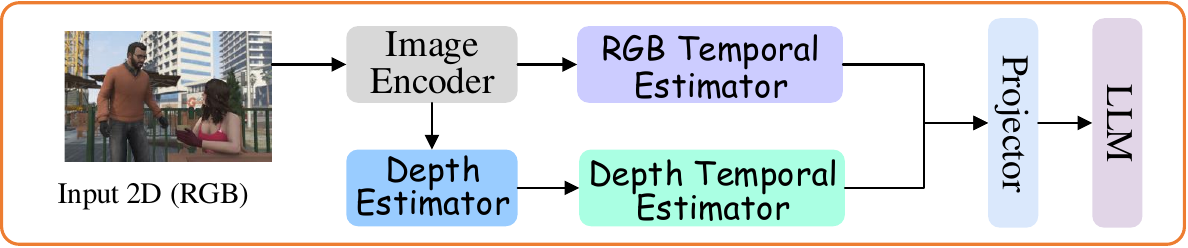}
    \caption{The inference stage of overall scene transcending (refer to step 4 in Fig. \ref{fig:frame}), where the input is only 2D (RGB) without 3D, video and 4D data.}
   \label{fig:2d-to-4d-inference}
\end{subfigure}
\vspace{-8mm}
\caption{Illustrations of 2D-to-4D visual scene transfer learning.}
\vspace{-6mm}
\label{fig:2D-TO-4D}
\end{figure}

In this section, we describe how to leverage large volumes of 2D scene data to aid 4D-PSG generation with limited annotated data. 
The core idea is that both 2D and 4D data can fundamentally describe visual scenes, which makes it possible to transfer the dimension-consistent scene-related knowledge embedded within these data.
Yet if relying solely on 2D data, we observe that 2D scenes lack the spatial depth and temporal dimensions inherent in 4D scenes, which limits the model's spatial and temporal awareness.
To address this, we introduce a spatial-temporal 2D-to-4D visual scene transfer learning D\textsuperscript{2$\to$4}-VST approach.
As illustrated in Fig. \ref{fig:2d-to-4d-overall}, overall, the training process consists of four key steps: \textbf{1): 4D Scene Perception Initiation Learning}, \textbf{2): 2D-to-4D Scene Transcending Learning}, \textbf{3): Pseudo 4D scene Transfer Initiation Learning}, and \textbf{4): Large-scale Visual Scene Transfer Learning}. 
To facilitate understanding, we also summarize all the used data sources and volumes at each step in Tab. \ref{tab:data_source}.

\vspace{-3mm}
\paragraph{$\blacktriangleright$ Step 1: 4D Scene Perception Initiation Learning}
\noindent\\
We begin with performing the initiation learning to enable the LLM to develop a foundational perception of the 4D scene so as to generate 4D SGs.
We employ the standard text generation loss $\mathcal{L}_{txt}$ for the optimization of textual SG triplets.
For tube masks, following \cite{next-chat-Zhang000C24}, we adopt IoU loss, Dice loss \cite{abs-2312-05391}, and Focal loss \cite{LinGGHD17} to optimize the model:
\begin{equation}
\label{eq:step-1}
    \mathcal{L}_{4dpsg} = \mathcal{L}_{txt} + \mathcal{L}_{IoU} + \mathcal{L}_{Dice} + \mathcal{L}_{Focal}.
\end{equation}

\begin{table}[!t]
\fontsize{8}{9}\selectfont
\setlength{\tabcolsep}{0.8mm}
\centering
\begin{tabular}{ccp{4.2cm}}
\toprule
\bf Step & \bf SG Learning & \bf 2D$\to$4D Scene Transcending \\
\midrule
\rowcolor{lightlightgrey}  \#1 & PSG4D \cite{4d-psg-yang20244d} (4D-SGs $|$ \textcolor{red}{3K}) & \multicolumn{1}{c}{\rule{0pt}{2ex} /} \\
\multirow{3}{*}{ \#2} & \multirow{3}{*}{/} &  
 DIML \cite{DIML-abs-2110-11590} (2D, 3D $|$ \textcolor{red}{200K}) \\
& & AG \cite{AG-JiK0N20} (Video $|$ \textcolor{red}{288K}) \\
& & PSG4D \cite{4d-psg-yang20244d} (Depth $|$ \textcolor{red}{3K}) \\
\rowcolor{lightlightgrey}  \#3 & \rule{0pt}{2ex}  PSG4D \cite{4d-psg-yang20244d} (4D-SGs $|$ \textcolor{red}{3K})  & PSG4D \cite{4d-psg-yang20244d} (2D, 3D, Vid, Dep $|$ \textcolor{red}{3K}) \\
 \#4 & \multicolumn{2}{c}{VG \cite{VG-KrishnaZGJHKCKL17}, PSG \cite{PSG-YangAGZZ022} (2D-SGs; 2D $|$ \textcolor{red}{150K})}  \\
\bottomrule
\end{tabular}
\vspace{-2mm}
\caption{The data sources and \textcolor{red}{data volume} used at each learning step for 4D SG generation, and 2D-to-4D Scene Transcending. 
In step 2, external data is used to optimize each estimator.
3D, Video (Vid.), and Depth (sequence) are all derived from PSG4D in step 3. 
}
\vspace{-5mm}
\label{tab:data_source}
\end{table}

\vspace{-4mm}
\paragraph{$\blacktriangleright$ Step 2: 2D-to-4D Scene Transcending Learning}
\noindent\\
As shown in Fig. \ref{fig:2d-to-4d-overall}, the learning of this step is divided into three sub-process: 
\textit{a) 2D (RGB) to Depth Transcending}, aiming to optimize the depth estimator $F_{de}$ to predict the depth features;
\textit{b) 2D (RGB) Temporal Transcending}, designed to generate a 2D (RGB) temporal sequence features using an RGB Temporal Estimator $F_{rte}$; 
\textit{c) Depth Temporal Transcending}, focusing on training a Depth Temporal Estimator $F_{dte}$ to yield depth temporal sequence features. 
Finally, the 2D (RGB) sequence features are combined with the depth temporal sequence features over time to form the 4D scene features. 
The learning process of each subprocess is detailed below.
Fig. \ref{fig:2D-TO-4D} illustrates all the mechanisms.

\indent \textbf{Subprocess-a): 2D (RGB) to Depth Transcending Learning.}
Given the input 2D RGB image $\mathcal{I}^{RGB}$ extracted from a 3D scene, the objective of the Depth Estimator $F_{de}$ is to infer accurate depth features. 
To achieve this, we employ ImageBind \cite{imagebind-GirdharELSAJM23} as the image encoder to process the input image and yield 2D scene features $\bm{H}^{RGB}$. 
Inspired by DPT \cite{DPT-RanftlBK21}, we design the Depth Estimator consisting of a convolutional layer and a projector to predict depth features. 
A regression loss is then computed between the predicted depth features and the ground-truth depth features $\bm{H}^{D}$ of 3D, obtained from a depth encoder, also implemented by ImageBind:
\setlength\abovedisplayskip{3pt}
\setlength\belowdisplayskip{3pt}
\begin{equation}
\label{eq:de}
    \mathcal{L}_{de} = \rho(F_{de}(\bm{H}^{RGB}), \bm{H}^{D}).
\end{equation}

\indent \textbf{Subprocess-b): 2D (RGB) Temporal Transcending Learning.}
Given an input 2D image scene $\mathcal{I}^{RGB}$ extracted from a video, the goal of the RGB Temporal Estimator $F_{rte}$ is to predict temporal features relevant to the current input. 
Following the approach in \cite{LiPKCFCV22}, we implement an autoregressive transformer that processes 2D image scene features to generate predicted temporal features, denoted as $\hat{\bm{H}}^{T}$:
\setlength\abovedisplayskip{3pt}
\setlength\belowdisplayskip{3pt}
\begin{equation}
\begin{aligned}
\label{eq:rte}
    p(\hat{\bm{H}}^{T}, \bm{H}^{RGB}, F_{rte}) = \prod_{j}F_{rte}(\hat{\bm{H}}^{T}_{ j}|\hat{\bm{H}}^{T}_{\le j}, \bm{H}^{RGB}).
\end{aligned}
\end{equation}
Similarly, the estimator is trained to minimize the regression loss between predicted features $\hat{\bm{H}}^{T}$ and gold features $\bm{H}^{T}$ obtained by a video encoder:
 \begin{equation}
     \mathcal{L}_{rte} = \rho(\hat{\bm{H}}^{T}, \bm{H}^{T}).
 \end{equation}

\indent \textbf{Subprocess-c): Depth Temporal Transcending Learning.}
Given the input depth $\mathcal{I}^{D}$ extracted from the depth sequence of a 4D scene, the Depth Temporal Estimator $F_{dte}$ adopts the same structure as the RGB Temporal Estimator, initializing it with $F_{rts}$ to ensure the stable generation of depth temporal features.
The regression loss is computed between the generated depth temporal features $\hat{\bm{H}}^{DT}$ and ground-truth $\bm{H}^{DT}$ to optimize $F_{dte}$: 
 \begin{equation}
 \label{eq:dte}
 \setlength\abovedisplayskip{1pt}
\setlength\belowdisplayskip{1pt}
     \mathcal{L}_{dte} = \rho(\hat{\bm{H}}^{DT}, \bm{H}^{DT}).
 \end{equation}

\vspace{-4mm}
\paragraph{$\blacktriangleright$ Step 3: Pseudo 4D Scene Transfer Initiation Learning}
\noindent\\
Upon completing the 2D-to-4D scene transcending learning, we achieve the pseudo 4D scene feature injection. 
Next, we use a limited amount of 4D data to further refine the transcending module, and also to directly apply the transcended 2D scene features into the full 4D-LLM to interpret and produce the 4D PSG.
In this learning phase, we retain the learning approach from Step 2. 
Also, to ensure consistency between the 2D (RGB) temporal sequence and the depth temporal sequence, we enforce that the predicted 2D features, generated by the Depth Estimator, closely match both the predicted and ground-truth depth features:
\setlength\abovedisplayskip{3pt}
\setlength\belowdisplayskip{3pt}
\begin{equation}
\begin{aligned}
\mathcal{L}_{dep}^{\heartsuit} &= \rho(F_{de}(F_{rts}(\bm{H}^{RGB})), \bm{H}^{D}), \\
\mathcal{L}_{dep}^{\diamondsuit} &= \rho(F_{dte}(F_{de}(\bm{H}^{RGB})), F_{de}(F_{rte}(\bm{H}^{RGB}))). \\
\end{aligned}
\end{equation}
Overall, the training loss can be formulated:
\setlength\abovedisplayskip{2pt}
\setlength\belowdisplayskip{2pt}
\begin{equation}
\label{eq:step-3}
\mathcal{L} = \mathcal{L}_{4dpsg} + \mathcal{L}_{de} + \mathcal{L}_{rte} + \mathcal{L}_{dte} + \mathcal{L}_{dep}^{\heartsuit} + \mathcal{L}_{dep}^{\diamondsuit}.
\end{equation}

\vspace{-3mm}
\paragraph{$\blacktriangleright$ Step 4: Large-scale Visual Scene Transfer Learning}
\noindent\\
Following scene transfer initiation learning, we leverage large volumes of 2D visual features (i.e., 2D SGs) to enhance 4D scene understanding for 4D-PSG generation. 
Specifically, the model takes only 2D scenes as input, which are then transcended into pseudo-4D scenes (cf. Fig. \ref{fig:2d-to-4d-inference}).
Then, the LLM generates the corresponding SG triplets, while the mask decoder produces segmentation masks for each object. 
For optimization, we use the gold 2D image SG as a supervision signal and apply the same loss as in Step 1.

\begin{table*}[!t]
\fontsize{8}{9}\selectfont
\setlength{\tabcolsep}{4.2mm}
\centering
\begin{tabular}{lcccccc}
    \toprule
    \multirow{2}{*}{\bf Method} & \multicolumn{3}{c}{\bf PSG4D-GTA} & \multicolumn{3}{c}{\bf PSG4D-HOI} \\
    \cmidrule(lr){2-4} \cmidrule(lr){5-7}
     & \bf R/mR@20 & \bf R/mR@50 & \bf R/mR@100 & \bf R/mR@20 & \bf R/mR@50 & \bf R/mR@100 \\
    \midrule
    3DSGG \cite{3DDSG-WaldDNT20} & 2.29 / 0.92 & 2.46 / 1.01 & 3.81 / 1.45 & 4.23 / 2.19 & 4.47 / 2.31 & 4.86 / 2.41 \\
    PSG4DFormer\textsuperscript{w/o t}~\cite{4d-psg-yang20244d} & 4.43 / 1.34 & 4.89 / 2.42 & 5.26 / 2.83 & 4.44 / 2.37 & 4.83 / 2.43 & 5.21 / 2.84 \\
    PSG4DFormer\textsuperscript{w/o d}~\cite{4d-psg-yang20244d} & 4.40 / 1.42 & 4.91 / 1.93 & 5.49 / 2.27 & 5.49 / 3.42 & 5.97 / 3.92 & 6.43 / 4.21 \\
    PSG4DFormer~\cite{4d-psg-yang20244d} & 6.68 / 3.31 & 7.17 / 3.85 & 7.22 / 4.02 & 5.62 / 3.65 & 6.16 / 4.16 & 6.28 / 4.97 \\
    PSG4DFormer\textsuperscript{one-stage} & 7.93 / 3.69 & 8.43 / 3.98 & 9.07 / 4.34  & 7.05 / 3.96  & 8.04 / 4.75  &  8.59 / 5.01 \\
    \cdashline{1-7}
    4D-LLM (ours, base) & 10.03 / 5.04  & 11.32 / 5.45 & 12.31 / 6.32 & 10.58 / 4.31 & 13.27 / 6.37 & 15.14 / 6.76 \\     
    \rowcolor{lightgrey} 4D-LLM (ours) & 18.48 / 9.43 & 20.24 / 10.78 & 18.43 / 12.41 & 15.18 / 8.02 & 19.68 / 10.23 & 21.79 / 11.54 \\
    \bottomrule
\end{tabular}
\vspace{-3mm}
\caption{Performance comparison across different methods on PSG4D-GTA and PSG4D-HOI datasets.``w/o t'' and ``w/o d'' denote the removal of the temporal encoder and depth branch, respectively. ``4D-LLM (our, base)'' means the model only is trained in step 1 with PSG4D data, without 2D scene enhancement.}
\label{tab:main_performance}
\vspace{-2mm}
\end{table*}

\begin{table*}[!t]
\fontsize{8}{9}\selectfont
\setlength{\tabcolsep}{4.3mm}
\centering
\begin{tabular}{lcccccc}
    \toprule
    \multirow{2}{*}{\bf Method} & \multicolumn{3}{c}{\bf PSG4D-GTA} & \multicolumn{3}{c}{\bf PSG4D-HOI} \\
    \cmidrule(lr){2-4} \cmidrule(lr){5-7}
     & \bf R/mR@20 & \bf R/mR@50 & \bf R/mR@100 & \bf R/mR@20 & \bf R/mR@50 & \bf R/mR@100 \\
    \midrule
    \rowcolor{lightgrey} 4D-LLM (ours) &18.48 / 9.43 & 20.24 / 10.78 & 18.43 / 12.41 & 15.18 / 8.02 & 19.68 / 10.23 & 20.79 / 11.54 \\
    \quad w/o $V^{2\rightarrow4}$-{VST} & 12.03 / 6.04  & 13.32 / 6.45 & 14.31 / 6.78 & 10.58 / 4.31 & 14.47 / 7.05 & 15.14 / 6.76  \\ 
    \quad w/o step 2 & 15.35 / 7.10 & 17.45 / 7.02 & 19.73 / 9.34 & 12.35 / 6.79 & 16.23 / 8.87 & 18.45 / 7.58 \\
    \quad w/o step 3 & 16.42 / 7.69 & 16.53 / 8.34 & 21.35 / 10.36 & 14.21 / 6.40 & 18.34 / 9.07 & 21.47 / 8.98 \\
    \quad w/o CI & 17.32 / 8.73 & 19.04 / 9.78 & 22.46 / 11.07 & 15.37 / 7.45 & 19.12 / 9.48 & 20.13 / 10.25 \\
    \bottomrule
\end{tabular}
\vspace{-3mm}
\caption{Ablation Study. 
``w/o $V^{2\rightarrow4}$-{VST}'' denotes we only train the model with PSG4D data without 2D-to-4D visual scene transfer learning, i.e., obtaining 4D-LLM (base). 
``w/o CI'' means LLM directly outputs the SG triplets without chained inference.}
\label{tab:ablation_study}
\vspace{-4mm}
\end{table*}

\vspace{-1mm}
\section{Experiments}

\vspace{-1mm}
\subsection{Settings}
\vspace{-1mm}

\paragraph{Datasets and Evaluation Metrics.} We employ two PSG4D \cite{4d-psg-yang20244d} datasets: \textbf{PSG4D-GTA} contains 67 synthetic videos in a third-person perspective, comprising 35 object categories, 43 relation categories, and 728 triplets. 
\textbf{PSG4D-HOI} includes 2,973 videos compiled from an egocentric perspective, including 46 object categories, 15 object-object relation categories, and 29,375 triplets.
We leverage large-scale (150k) 2D scenes datasets, i.e., \textbf{VG} \cite{VG-KrishnaZGJHKCKL17} and \textbf{PVSG} \cite{PVSG-YangPLGCL0ZZLL23}, to enhance perception learning. 
Following \cite{4d-psg-yang20244d}, we evaluate the performance on SG generation tasks, and report the R@$K$ and mR@$K$ metrics, where $K=20, 50, 100$.

\begin{table}[!t]
\fontsize{8}{9}\selectfont
\setlength{\tabcolsep}{1.7mm}
\centering
\begin{tabular}{lcccc}
    \toprule
    \multirow{2}{*}{} & \bf Object & \bf Object Pair & \bf Triplets & \bf Quintuple \\
    &  (stage-1) & (stage-2) & (stage-3) & (stage-4) \\
    \midrule
    $\bullet$ PSG4D-GTA \\
     \quad w/ CI & 78.35  & 46.40  &  34.27  & 18.32   \\
     \quad w/o CI & 65.21 &  30.03 & 16.38 & 8.73\\ 
     \cdashline{1-5}
     $\bullet$ PSG4D-HOI \\
     \quad w/ CI & 75.24 & 60.09 & 45.03  & 16.78 \\
     \quad w/o CI & 61.06 &  40.36 & 20.89 & 8.91 \\ 
    \bottomrule
\end{tabular}
\vspace{-2mm}
\caption{The performance (R@20) on PSG4D-GTA and PSG4D-HOI at each chained inference (CI) stage.
}
\label{tab:CI_analysis}
\vspace{-5mm}
\end{table}

\vspace{-3mm}
\paragraph{Implementation.}
We employ Imagebind \cite{imagebind-GirdharELSAJM23} as our 4D scene encoder and, similarly, the image encoder, video decoder, and depth encoder when performing the 2D-to-4D transfer learning.
The projector layer and depth estimator are both implemented as a 2-layer MLP. 
Both the RGB Temporal Estimator and the Depth Temporal Estimator use 6 transformer layers with a 512-dimensional embedding dimension and 8 attention heads. 
The LLM is instantiated with LLaMA2 \cite{llama2-abs-2307-09288} and fine-tuned using LoRA \cite{lora-HuSWALWWC22}. 
We initialize the mask decoder with SAM2 \cite{sam2-abs-2408-00714} weights. 
The optimizer is AdamW, with an inverse square root learning rate schedule and warm-up steps. 
The training is carried out end-to-end on 8 H100 80GB GPUs.
We detail the learning parameters and architecture implementation in the Appendix.

\subsection{Main Results}

We first evaluate our proposed method against state-of-the-art (SoTA) baselines on the PSG4D dataset, with results presented in Tab. \ref{tab:main_performance}.
In contrast to pipeline-based baseline approaches, we also implement an end-to-end model for 4D-PSG generation. 
The results indicate consistent performance improvements on two PSG4D datasets, suggesting that the end-to-end approach effectively mitigates error propagation to improve performance.
Furthermore, we observe that the LLM-based baseline, even when trained on a limited amount of PSG4d data, yields substantial performance gains, with average improvements of 4.74 mR@20 score on PSG4D-GTA and 5.50 on PSG4D-HOI compared to traditional specialist models. 
This enhancement is likely contributed by the LLM's powerful capabilities in understanding and reasoning.
Our approach, which integrates 2D-to-4D scene transfer learning, achieves further performance improvements, establishes a new SoTA, and demonstrates the effectiveness of scene enhancement learning in 4D-PSG tasks.

\begin{table}[t]
\fontsize{8}{9}\selectfont
\setlength{\tabcolsep}{0.45mm}
\centering
\begin{tabular}{llcccc}
    \toprule
    \multirow{2}{*}{\bf Modality} & \multirow{2}{*}{\bf Data} & \multicolumn{2}{c}{\bf PSG4D-GTA} & \multicolumn{2}{c}{\bf PSG4D-HOI} \\
    \cmidrule(lr){3-4} \cmidrule(lr){5-6}
     & & \bf R/mR@20 & \bf R/mR@50 & \bf R/mR@20 & \bf R/mR@50 \\
    \midrule
     2D & VG \cite{VG-KrishnaZGJHKCKL17} &  14.24 / 6.95 & 15.25 / 7.02 & 11.76 / 5.01 & 15.86 / 8.96  \\
     Video & AG \cite{AG-JiK0N20} & 12.34 / 6.45 & 14.38 / 6.49 & 10.54 / 4.86 & 15.31 / 8.47\\
    3D & 3DDSG \cite{3DDSG-WaldDNT20} & 12.07 / 5.47 & 13.56 /6.02  & 12.35 / 4.58 & 15.30 / 7.08 \\
    \bottomrule
\end{tabular}
\vspace{-3mm}
\caption{Comparison of enhancing the model's 4D scene perception capability by directly using SG data from different modalities.
}
\label{tab:direct_use}
\vspace{-2mm}
\end{table}

\begin{table}[!t]
\fontsize{8}{9}\selectfont
\setlength{\tabcolsep}{0.5mm}
\centering
\begin{tabular}{ccccccc}
    \toprule
    \multicolumn{3}{c}{\bf Data Source} & \multicolumn{2}{c}{\bf PSG4D-GTA} & \multicolumn{2}{c}{\bf PSG4D-HOI} \\
    \cmidrule(lr){1-3} \cmidrule(lr){4-5} \cmidrule(lr){6-7}
    \bf Image & \bf Video & \bf 3D & \bf R/mR@20 & \bf R/mR@50 & \bf R/mR@20 & \bf R/mR@50  \\
    \midrule
    \checkmark & $\times$ & $\times$ & 18.48 / 9.43 & 20.24 / 10.78 & 15.18 / 8.02 & 19.68 / 10.23 \\
    $\times$ & \checkmark & $\times$ & 16.83 / 7.98 & 17.89 / 8 .50  & 13.94 / 7.12 & 17.03 / 9.07 \\
    $\times$ & $\times$ & \checkmark & 12.78 / 6.96 & 13.68 / 6.76 & 10.75 / 4.36 &  16.32 / 8.45  \\
     
    \bottomrule
\end{tabular}
\vspace{-3mm}
\caption{
Comparison between using different data sources at step 4 for enhancing 4D scene perception.
}
\label{tab:diff_data}
\vspace{-4mm}
\end{table}

\vspace{-1mm}
\subsection{System Ablation Study}

\vspace{-1mm}
Next, we conduct an ablation study to examine the contribution and necessity of each component within the proposed framework, as shown in Tab. \ref{tab:ablation_study}.
Firstly, we remove the 2D-to-4D visual scene transfer learning and observe the most significant drop in model performance.
We then analyze the individual impacts of steps 2 and 3.
Removing step 3 results in a performance decline, likely due to the loss of additional alignment learning that step 3 provides for maintaining scene consistency across pseudo 4D scene data.
In comparison, omitting step 2 also leads to a marked performance decrease, suggesting that step 2 plays a foundational role in scene transcending. 
Without it, alignment learning solely on limited 4D data proved insufficient, which is further substantiated by subsequent experiments in RQ-2.
Finally, we examine the effect of chained inference, noting a performance decrease when removed, which highlights its effectiveness. Further in-depth analyses are provided below.

\begin{table}[t]
\fontsize{8}{9}\selectfont
\setlength{\tabcolsep}{1.6mm}
\centering
\begin{tabular}{lcccc}
    \toprule
    \multirow{2}{*}{\bf Method} & \multicolumn{2}{c}{\bf Seen} & \multicolumn{2}{c}{\bf Unseen} \\
    \cmidrule(lr){2-3} \cmidrule(lr){4-5}
     & \bf Object & \bf Predicate & \bf Object & \bf Predicate \\
    \midrule
    PSG4DFormer & 40.02  & 12.68  & 0.47 & 0.31  \\
    4D-LLM (base) & 47.32 & 25.67  & 27.36  & 20.12  \\
    4D-LLM (base) w/ CI & 48.70  & 29.24 & 34.15 &  26.86 \\
    4D-LLM  & 60.59 & 40.78  & 45.24 & 27.96 \\
    \bottomrule
\end{tabular}
\vspace{-2mm}
\caption{Performance comparison for seen and unseen objects, as well as seen and unseen relations. ``w/ CI'' indicates the employment of chained inference.
}
\label{tab:open_vocab}
\vspace{-4mm}
\end{table}

\vspace{-2mm}
\section{Analyses and Discussions}
\vspace{-1mm}

In this section, we give more discussions via a series of
in-depth analyses to reveal how the system advances.
Following, we try to ground the answers for the following
five key research questions.

\vspace{-3mm}
\paragraph{RQ-1: Is the 2D-to-4D scene transcending necessary?}

In the ablation study, we demonstrated the benefits of transfer learning, which led to significant performance improvements.
Here, we aim to further investigate whether 2D-to-4D scene transcending is essential, or if directly using 2D SG data can enhance 4D-PSG understanding without transcending.
To explore this, we directly encode 2D image scenes using the RGB encoding module in the 4D scene encoder and feed them into the 4D-LLM for learning. 
The results, presented in Tab. \ref{tab:direct_use}, indicate that using 2D scenes with transcending yields a performance improvement over the baseline 4D-LLM, suggesting that 2D scenes provide valuable scene knowledge that enhances 4D scene understanding.
However, as shown in Tab. \ref{tab:ablation_study}, without Step 2 and relying on limited 4D scene transcending, directly learning from 2D scenes remains suboptimal, underscoring the necessity of scene transcending. 
Additionally, we conduct a comparative analysis with other data modalities used for enhancement. 
While these also improve performance, they fall short compared to our model that employs 2D-to-4D scene transcending.

\begin{table}[t]
\fontsize{8}{10}\selectfont
\setlength{\tabcolsep}{1.3mm}
\centering
\begin{tabular}{lcccccc}
    \toprule
    \multirow{2}{*}{\bf Method} & \multicolumn{3}{c}{\bf Object} & \multicolumn{3}{c}{\bf Predicate} \\
    \cmidrule(lr){2-4} \cmidrule(lr){5-7}
     & \bf Head &\bf  Body & \bf Tail &\bf  Head & \bf Body &\bf  Tail \\
    \midrule
    $\bullet$ \bf PSG4D-GTA \\
     \quad PSG4DFormer & 89.48 & 66.79 & 23.68 & 78.98 & 65.53 & 10.39 \\
     \quad 4D-LLM (base) & 90.13 & 73.41 & 48.23 & 83.40 & 70.56 &  33.56 \\
     \quad 4D-LLM (ours) & 98.78 & 87.28 & 61.43  & 90.57  & 82.10 & 55.48 \\
     \hline
     $\bullet$ \bf PSG4D-HOI \\
     \quad PSG4DFormer & 73.41 & 45.34 & 18.65 &  89.59 & 70.53 & 28.6 \\
     \quad 4D-LLM (base) & 85.03 & 65.30 & 38.59 &  93.33 & 84.56 &  59.7\\
     \quad 4D-LLM (ours) & 91.36 & 78.44 & 51.48 &  99.86 & 89.67 &  83.6 \\
    \bottomrule
\end{tabular}
\vspace{-2mm}
\caption{Performance comparison on head, body, and tail objects and Predicates recognition.}
\label{tab:head_body_tail}
\vspace{-2mm}
\end{table}

\begin{figure}[!t]
    \centering
    \begin{subfigure}{0.46\linewidth}
        \centering
        \includegraphics[width=\linewidth]{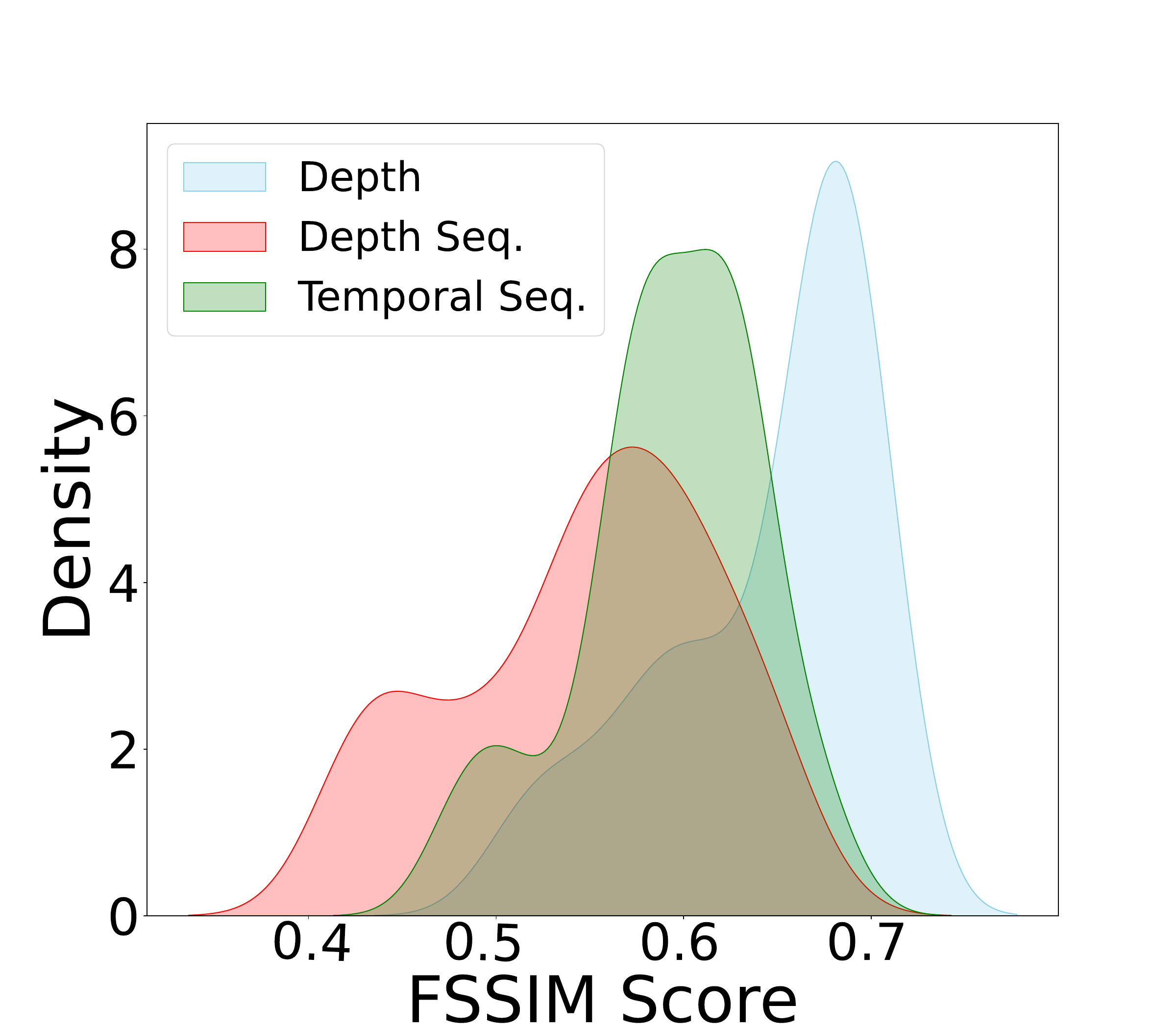}
        \vspace{-5mm}
        \caption{ }
        \label{fig:subfig1}
    \end{subfigure}
    \begin{subfigure}{0.46\linewidth}
        \centering
        \includegraphics[width=\linewidth]{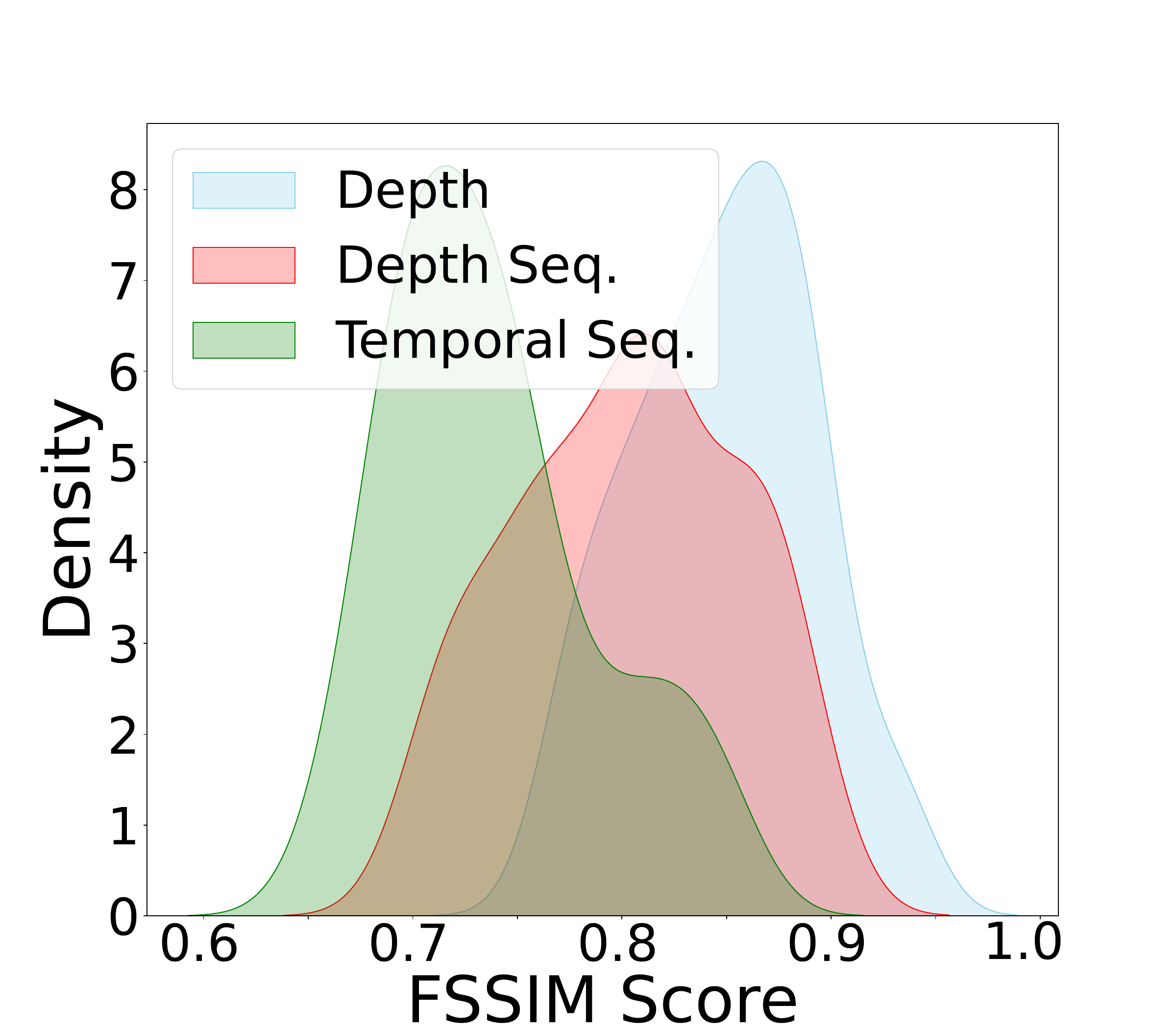}
        \vspace{-5mm}
        \caption{ }
        \label{fig:subfig2}
    \end{subfigure}
    \vspace{-4mm}
    \caption{The feature similarity distribution between predicted and gold ones without (a) and with (b) step 2.}
    \label{fig:feature_sim}
    \vspace{-4mm}
\end{figure}

\vspace{-3mm}
\paragraph{RQ-2: How effective is the quality of 2D-to-4D scene transcending?}
Here, we investigate whether the 2D-to-4D scene transcending process effectively learns spatial and temporal priors. 
To evaluate this, we analyze the learned depth and temporal sequence features by calculating their similarity \cite{Kobayashi16} with gold standard features and performing a visual analysis of the results, as depicted in Fig. \ref{fig:feature_sim}. 
Higher similarity values indicate closer alignment with ground truth.
Our findings reveal a high similarity between the model's predicted features and the ground truth, suggesting that the model accurately learns reasonable depth and temporal features.
Additionally, by comparing feature similarity with and without Step 2 learning, we observe that Step 2 significantly enhances feature similarity.
This largely contributes to improved 4D scene perception, as it enables the generation of pseudo-4D scenes closer to the actual 4D structure.
We further analyze whether using only certain modules in the scene transcending process yields performance improvements. 
For instance, when the input is a 3D scene, we use only the RGB Temporal Estimator and Depth Temporal Estimator. 
As shown in Tab. \ref{tab:diff_data}, our model still achieves performance gains compared to using only 3D data (see Tab. \ref{tab:direct_use}). 
This demonstrates the effectiveness of each module within our scene-transcending framework.

\begin{figure}[!t]
    \centering
    \includegraphics[width=0.98\linewidth]{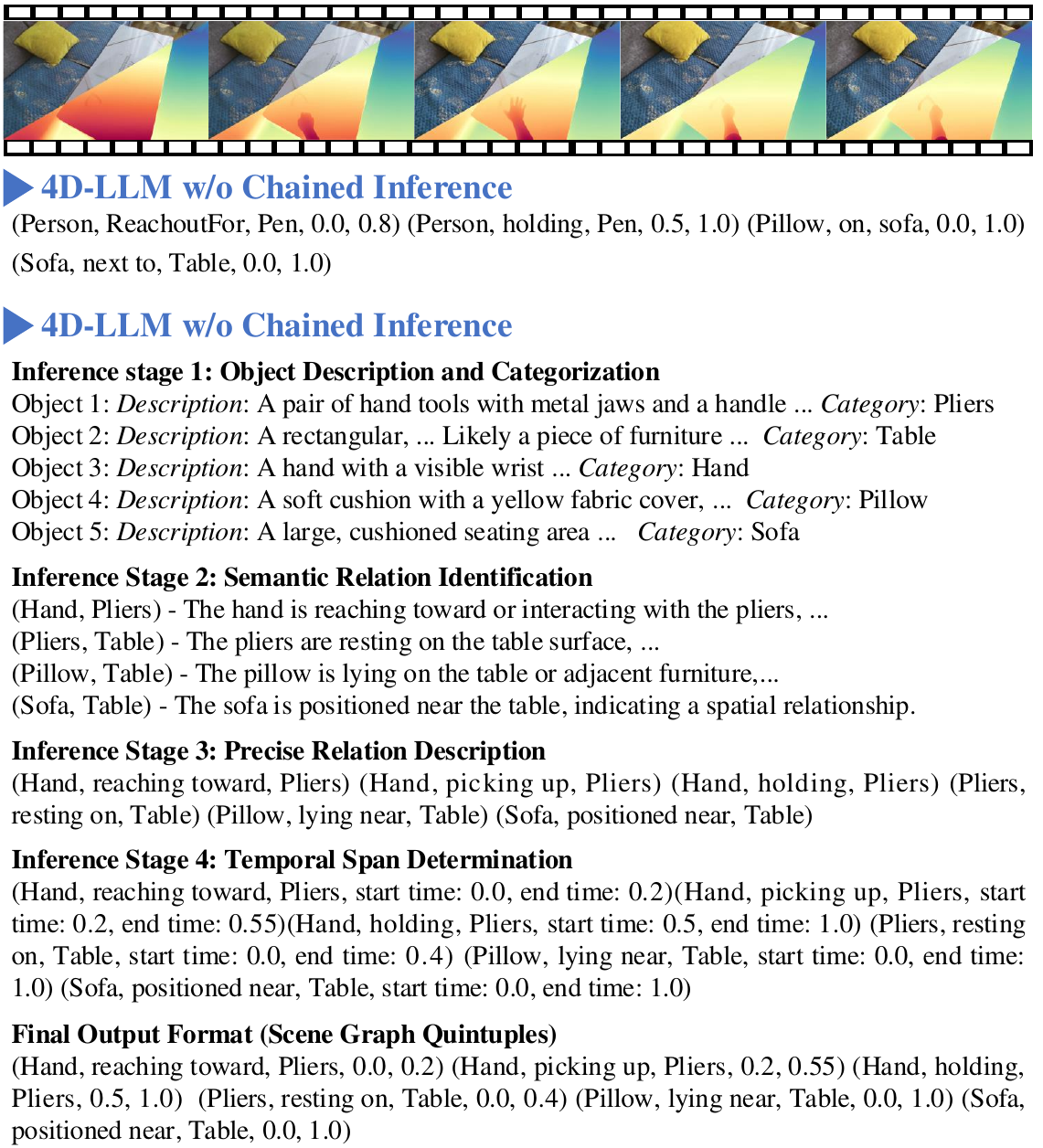}
    \vspace{-3mm}
    \caption{An instance for comparing 4D-LLM with/without chained SG inference mechanism.}
    \label{fig:CI_visualization}
    \vspace{-4mm}
\end{figure}

\vspace{-3mm}
\paragraph{RQ-3: How exactly are the advantages of the chained inference?}
To investigate this question, we first conduct a comparative analysis of the results at each stage of the chained inference process. The analysis results are shown in Tab. \ref{tab:CI_analysis}.
When chained inference is not applied, we calculate the performance of each subpart based solely on the final output.
The results indicate that, with the integration of chained inference, our model achieves better performance in each subpart, including object detection, object pair identification, predicate classification, and time span determination, compared to single-pass inference. 
This demonstrates that chained inference, by simplifying the reasoning process at each stage, can progressively enhance task performance.
Additionally, we visualize the model outputs with and without chained inference in Fig. \ref{fig:CI_visualization}. 
The visualization illustrates that, with chained inference, the model can analyze SG triplets step-by-step, leading to more accurate identification of objects and their relationships. 
This provides clear evidence of the effectiveness of chained inference.

\vspace{-2mm}
\paragraph{RQ-4: How well is the capability of solving open-vocabulary and zero-shot 4D-PSG?}
\noindent\\
\indent\textbf{1) Open-vocabulary Analysis.}
We present the model's performance on seen and unseen objects and predicates in Tab. \ref{tab:open_vocab}.
The results show that the baseline model lacks the capability for open-vocabulary settings, whereas the 4D-LLM baseline demonstrates some ability to recognize unseen objects and predicates, likely due to the extensive knowledge modeled within the LLM. 
With the integration of chained inference, the recognition capability of unseen categories of models improves significantly, as the chained inference helps guide the LLM through a step-by-step analysis to reach accurate results.
Additionally, a comparative analysis of performance across head, body, and tail objects and predicates reveals that our model consistently shows notable improvements on tail objects and predicates, further highlighting its effectiveness in handling less common categories.

\begin{figure}[ht]
    \centering
    \includegraphics[width=0.98\linewidth]{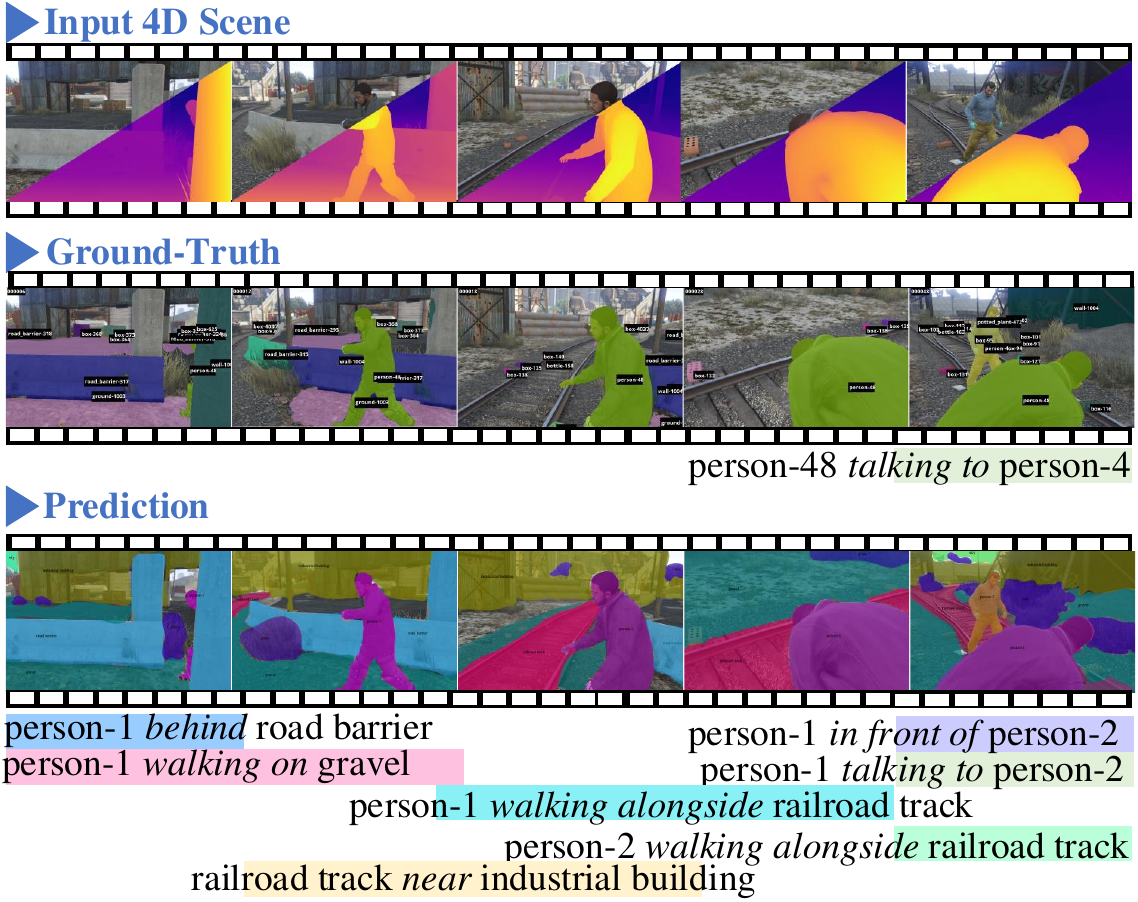}
    \vspace{-2mm}
    \caption{A case illustrating the prediction of 4D-LLM on 4D-PSG.}
    \label{fig:case}
    \vspace{-4mm}
\end{figure}

\indent\textbf{2) Zero-shot Study.}
We conduct a zero-shot study by utilizing only the 4D data for scene transcending learning, excluding PSG4D annotation data from training.
The results are presented in Appendix Tab. 3.
Our find that the baseline PSG4DFormer, which incorporates 2D-to-4D scene transfer learning on large-scale data, achieves performance comparable to supervised learning. 
This demonstrates that the visual knowledge embedded in 2D scenes can indeed transfer effectively and enhance 4D scene understanding. 
Furthermore, with the powerful reasoning capabilities of the LLM, the model's performance significantly improves, even surpassing that of the top-performing specialist.

\vspace{-2mm}
\paragraph{RQ-5: How exactly performance of 4D-LLM in 4D-PSG task?}

In Fig. \ref{fig:case}, we present a visual example of the predictions made by the 4D-LLM, offering a clear illustration of its capabilities.
Our model can identify more complex and detailed objects, such as industrial buildings and railroad tracks, and it describes relations between objects with high precision and flexibility, for instance, such as like ``person-1 walking alongside railroad track'', and ``person-1 walking on gravel''.
This level of detail is highly advantageous for downstream applications of SGs.
We offer more examples in the Appendix.

\vspace{-1mm}
\section{Conclusion}

\vspace{-0mm}
In this paper, we introduce a novel framework for 4D-PSG generation, effectively addressing the challenges of data scarcity issue. 
Our approach integrates a 4D-LLM with a 3D mask decoder as the backbone, enabling the end-to-end 4D-PSG generation. 
We design a chained SG inference mechanism that leverages the open-vocabulary capabilities of LLMs to infer accurate and detailed object and relation labels. 
Further, we develop a 2D-to-4D visual scene transfer learning strategy, which employs a scene transcending module to transfer modality-invariant features from abundant 2D SG annotations to 4D scenes, effectively compensating for data scarcity.
Experiments demonstrate that our system outperforms baselines significantly.
In-depth analyses further show the effectiveness of the chained inference mechanism, and scene transfer learning approach.

\clearpage

{
    \small
    \bibliographystyle{ieeenat_fullname}
    \bibliography{main}
}

\clearpage

\input{X_suppl}

\end{document}

%% file: preamble.tex
\usepackage{url}
\usepackage{amsfonts}       
\usepackage{nicefrac}       
\usepackage{microtype}      
\usepackage{xcolor}         
\usepackage{arydshln}
\usepackage[normalem]{ulem}
\usepackage{color}
\usepackage{graphicx}
\usepackage{amsmath,bm}
\usepackage{amssymb}
\usepackage{array}
\usepackage{comment}
\usepackage{booktabs}
\usepackage{caption}
\usepackage{colortbl} 
\usepackage{multirow}
\usepackage{makecell}
\usepackage{wrapfig}
\usepackage{pifont}
\usepackage{tabularray}
\usepackage{paralist}
\usepackage{adjustbox}
\usepackage{rotating}
\usepackage{enumerate}
\usepackage{enumitem}
\usepackage{CJKutf8}
\usepackage{algorithm}
\usepackage{algpseudocode}
\usepackage{tcolorbox}
\usepackage{adjustbox}
\usepackage{pgfplots}
\pgfplotsset{compat=1.17}
\tcbuselibrary{breakable}
\newcommand{\customfont}{\linespread{0.7}\selectfont}

\definecolor{myblue}{RGB}{52,218,247}
\definecolor{myred}{RGB}{255,90,90}
\definecolor{mypink}{RGB}{239,43,159}
\definecolor{myupdate}{RGB}{254,243,222}
\definecolor{myfrozen}{RGB}{237,255,255}
\definecolor{ired}{RGB}{229,72,72}
\definecolor{igreen}{RGB}{3,196,144}
\definecolor{nmblue}{RGB}{216,234,247}
\definecolor{lightgreen}{RGB}{196,250,222}
\definecolor{lightblue}{RGB}{211,227,253}
\definecolor{lightgrey}{RGB}{201,203,209}
\definecolor{lightlightgrey}{RGB}{240,240,240}
\definecolor{linkred}{RGB}{255,0,49}
\definecolor{textred}{RGB}{255,242,234}
\definecolor{imageblue}{RGB}{224,250,255}
\definecolor{videogreen}{RGB}{214,250,232}
\definecolor{vgreen}{RGB}{39,190,110}
\definecolor{linkcolor}{RGB}{255,0,0}
\definecolor{urlcolor}{RGB}{255,105,180}
\definecolor{citepcolor}{RGB}{66,168,235}
\definecolor{ipurple}{RGB}{126,3,196}
\definecolor{ired}{RGB}{224,0,0}

\newcommand{\circlenum}[1]{%
    \resizebox{!}{0.8em}{%
        \tikz[baseline=(char.base)]{
            \node[shape=circle, fill=black, inner sep=0.8pt, text=white] (char) {#1};
        }%
    }%
}

\newcommand{\logo}{\raisebox{-6pt}{\includegraphics[width=1.5em]{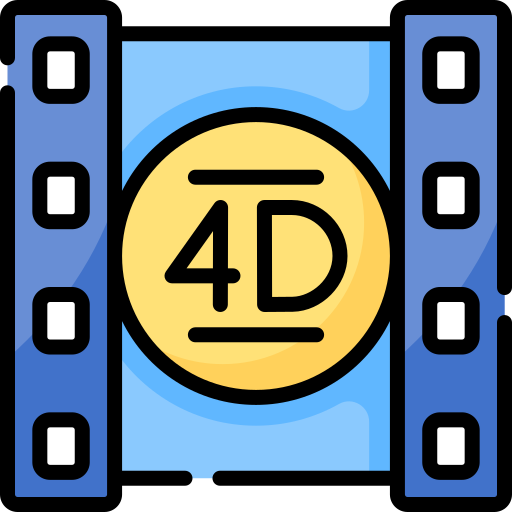}}\xspace\xspace\xspace}

%% file: X_suppl.tex



\clearpage
\setcounter{page}{1}
\maketitlesupplementary
\appendix

\section*{Overview}

The appendix presents more details and additional results not included in the main paper due to page limitation. The list of items included are:

\begin{compactitem}
    \item Potential Limitation and Future Work in $\S$\ref{app:limitation-future};
    \item Extended Task Definition in $\S$\ref{app:task-definition};
    \item Framework Architecture in $\S$\ref{app:framework};
    \item Detailed Training Procedure in $\S$\ref{app:training};
    \item Detailed System Inference in $\S$\ref{app:inference};
    \item Dataset Specification in $\S$\ref{app:datasets};
    \item Detailed Experimental Implementations in $\S$\ref{app:implementations};
    \item Additional Experiments in $\S$\ref{app:experiments}.
\end{compactitem}

\section{Potential Limitation and Future Work}
\label{app:limitation-future}

\subsection{Potential Limitations of 4D-LLM}

Despite its promising performance, 4D-LLM faces certain limitations. 
First, the dataset annotations used for training and evaluation may suffer from incomplete labeling or annotation errors, which may hinder the model's ability to learn accurate representations and achieve optimal performance. These limitations in the dataset may result in missed detections or incorrect relationships within the generated scene graphs, reducing the reliability of the model in practical applications.
The proposed approach, chained inference, which leverages the inherent reasoning capabilities of LLMs, has shown the potential to alleviate this issue by improving the consistency and robustness of the model’s predictions.

Another potential limitation is the model's capacity to handle extreme long-term 4D scenes. 
Understanding and processing extended temporal sequences in dynamic 4D environments remains challenging due to the complexity of capturing and reasoning about long-term dependencies and interactions. 
Current models, including 4D-LLM, may struggle to maintain coherence and accuracy in such scenarios, which is critical for tasks requiring an understanding of prolonged events or activities, such as surveillance or continuous monitoring.

\vspace{-2mm}
\subsection{Future Works}

Several directions can be pursued to enhance and expand the applications of 4D-LLM. 
One significant application area is robotics, where processing and understanding rich 4D scene information could greatly enhance robotic perception, decision-making, and task execution. For instance, by leveraging 4D-LLM, robots could gain a comprehensive understanding of their surrounding environments, enabling them to make informed decisions and adapt their actions in real time. This capability is particularly relevant for tasks such as autonomous navigation in dynamic and complex environments, where accurate scene understanding is essential for planning and executing tasks with high precision and safety.

Beyond robotics, 4D-LLM holds potential in virtual environments, such as serving as an autonomous agent in video games like GTA. Unlike traditional task completion systems that passively respond to predefined inputs, 4D-LLM could actively perceive and interpret its surroundings, dynamically interacting with the environment to achieve objectives. This transition from passive to active perception and decision-making highlights a shift toward greater autonomy and adaptability in AI systems.

\begin{figure}[!h]
    \centering
    \includegraphics[width=0.99\linewidth]{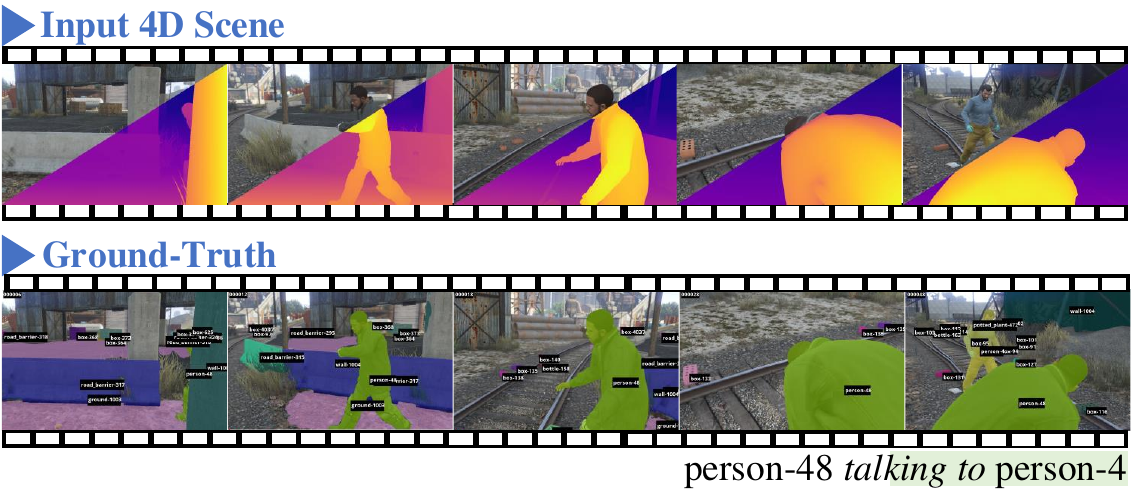}
    \vspace{-2mm}
    \caption{Input and output of the 4D panoptic scene graph (4D-PSG) generation task.}
    \label{fig:app-4dpsg}
    \vspace{-2mm}
\end{figure}

\vspace{-2mm}
\section{Extended Task Definition}
\label{app:task-definition}

As shown in Fig. \ref{fig:app-4dpsg}, given a 4D scene, specifically represented as a sequence of RGB-D frames $\mathcal{I} \in \mathbb{R}^{T \times H \times W \times 4}$, where $T$ denotes the number of frames and each frame has dimensions $H \times W \times 4$, our objective is to generate a dynamic panoptic SG $\mathcal{G} = \{\mathcal{O}, \mathcal{M}, \mathcal{R}\}$. 
The RGB-D sequence can also be treated as two parallel sequences: RGB images and single-channel depth images.
Here, $\mathcal{O} = \{\bm{o}_n\}_{n=1}^{N}$ are the set of objects present in the scene, for example, in Fig. \ref{fig:app-4dpsg}, the object, ``\textit{person-48}'', ``\textit{person-4}'', ``\textit{road-barrier-295}'', ``\textit{wall-1004}'', etc.  
$\mathcal{M} = \{\bm{m}_n\}_{n=1}^{N}$ denotes the corresponding binary mask tubes, where $\bm{m}_i \in \{0,1\}^{T \times H \times W}$ tracks the spatial extent of object $o_i$ over time $T$, as illustrated by the color-coded regions in Fig. \ref{fig:app-4dpsg}.
The relation set $\mathcal{R}=\{\bm{r}_k\}_{k=1}^{K}$ defines interactions between objects, with each $\bm{r}_k$ linking a subject and an object through a predicate class over a specific period $(t_s, t_e)$.
For instance, the relation, ``\textit{person-48 talking to person-4}'', is recognized, with its temporal duration represented by the length of the corresponding color block.

\begin{figure}
    \centering
    \includegraphics[width=0.99\linewidth]{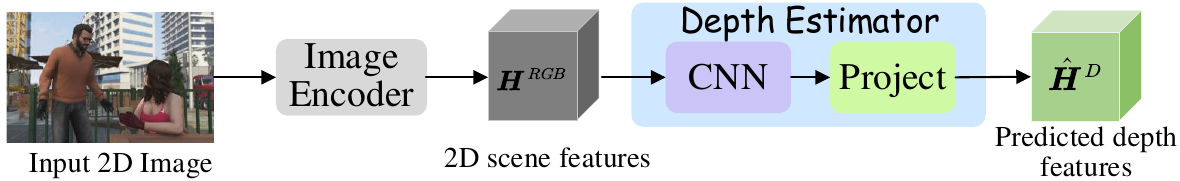}
    \vspace{-2mm}
    \caption{The framework of Depth Estimator.}
    \label{fig:depth-estimator}
\end{figure}

\vspace{-1mm}
\section{Framework Architecture}
\label{app:framework}
\vspace{-1mm}

Here, we detailed the framework architecture of the three estimators employed in the spatial-temporal 2D-to-4D transcending mechanism. 

\vspace{-3mm}
\paragraph{Depth Estimator.}
As depicted in Fig. \ref{fig:depth-estimator}, given an input 2D image $\mathcal{I} \in \mathbb{R}^{H \times W \times 3}$, an image encoder is to model the input image and yield 2D scene features $\bm{H}^{RGB} \in \mathbb{R}^{\frac{H}{p} \times \frac{W}{p} \times d }$, where $p$ represents the patch size, $d$ is the feature dimensionality.
Then, we implement the convolution (CNN) with a $3 \times 3$ kernel and then a projector using $1 \times 1 $ convolutions to project the input representation to match the dimension of ground-truth depth features $\bm{H}^{D}$.

\begin{figure}
    \centering
    \includegraphics[width=0.99\linewidth]{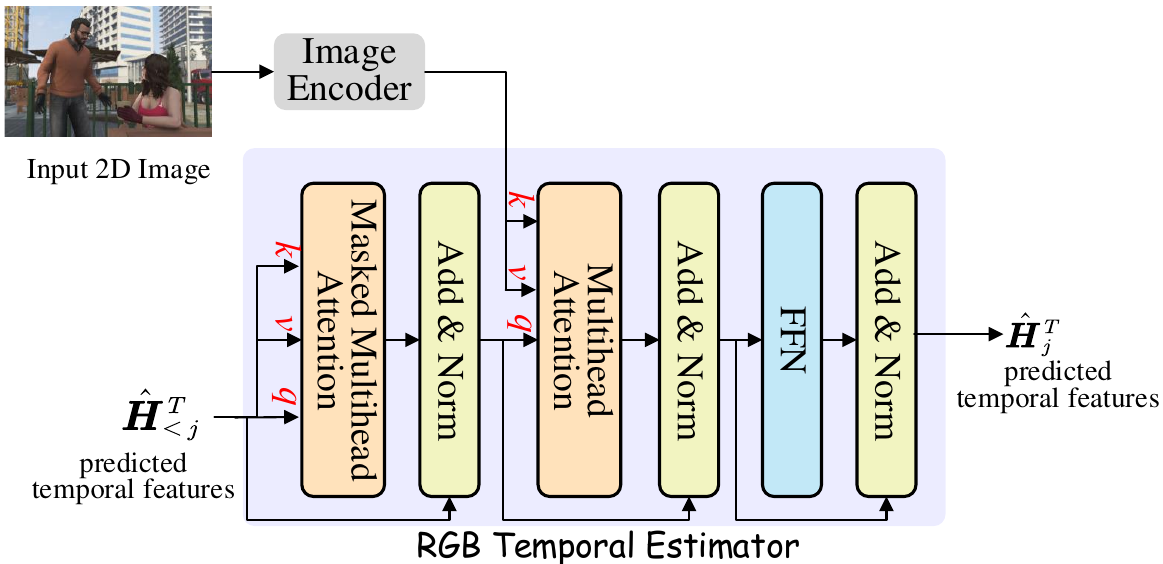}
    \vspace{-2mm}
    \caption{The framework of RGB Temporal Estimator.}
    \label{fig:RGB-temporal-estimator}
    \vspace{-2mm}
\end{figure}

\vspace{-3mm}
\paragraph{RGB Temporal Estimator.}

As shown in Fig. \ref{fig:RGB-temporal-estimator}, given an input 2D image $\mathcal{I} \in \mathbb{R}^{H \times W \times 3}$, an image encoder is to model the input image and yield 2D scene features $\bm{H}^{RGB} \in \mathbb{R}^{\frac{H}{p} \times \frac{W}{p} \times d }$, where $p$ represents the patch size, $d$ is the feature dimensionality. 
Inspired by \cite{transformer-VaswaniSPUJGKP17}, we design the RGB Temporal Estimator $F_{rte}$ as an autoregressive transformer, which is applied to predict the temporal features.
The masked multi-head attention, combined with the fact that the output features are offset by one position, ensures that the predictions for position $j$ can depend only on the known outputs at positions $< j$.
This autoregressive mechanism enables effective modeling of temporal dependencies. 
The probabilistic formulation is as follows:
\begin{equation}
\begin{aligned}
    p(\hat{\bm{H}}^{T}, \bm{H}^{RGB}, F_{rte}) = \prod_{j}F_{rte}(\hat{\bm{H}}^{T}_{ j}|\hat{\bm{H}}^{T}_{< j}, \bm{H}^{RGB}),
\end{aligned}
\end{equation}
where $\hat{\bm{H}}^{T}_{ j} $ denotes the predicted temporal features at step 
$j$, and $\hat{\bm{H}}^{T}_{<j} $ refers to the features predicted for previous steps.

\begin{figure}
    \centering
    \includegraphics[width=0.99\linewidth]{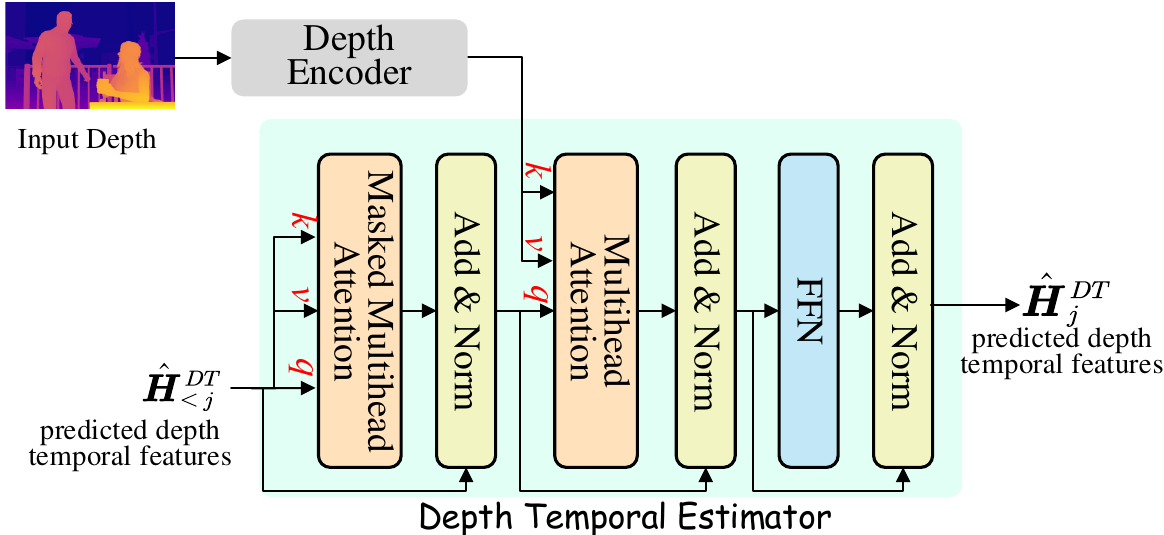}
    \vspace{-2mm}
    \caption{The framework of Depth Temporal Estimator.}
    \label{fig:depth-temporal-estimator}
    \vspace{-4mm}
\end{figure}

\vspace{-3mm}
\paragraph{Depth Temporal Estimator.}
As shown in Fig. \ref{fig:depth-temporal-estimator}, given an input depth image $\mathcal{I} \in \mathbb{R}^{H \times W \times 1}$, a depth encoder is to model the input image and yield depth features $\bm{H}^{D} \in \mathbb{R}^{\frac{H}{p} \times \frac{W}{p} \times d }$, where $p$ represents the patch size, $d$ is the feature dimensionality. 
Similarly, the Depth Temporal Estimator $F_{dte}$ employs the same architecture but different parameters to predict the depth temporal features:
\begin{equation}
\begin{aligned}
    p(\hat{\bm{H}}^{DT}, \bm{H}^{D}, F_{dte}) = \prod_{j}F_{dte}(\hat{\bm{H}}^{DT}_{ j}|\hat{\bm{H}}^{DT}_{< j}, \bm{H}^{D}).
\end{aligned}
\end{equation}

\vspace{-1mm}
\section{Extended Training Framework}
\label{app:training}
\vspace{-1mm}

This section details the training process, including the datasets, quantities, and parameters utilized at each step. 

\vspace{-2mm}
\paragraph{$\blacktriangleright$ Step 1: 4D Scene Perception Initiation Learning.}

The training begins with 4D Scene Perception Initiation Learning, designed to establish a foundational understanding of 4D scenes in the LLM for generating 4D-PSG.
In this step, we utilize the PSG4D dataset to train the 4D-LLM. 
The input comprises 4D scenes, while the supervision signals are the ground-truth 4D-PSGs, which include textual scene graph triplets and corresponding object mask tubes. 
The optimization is performed according to Eq.(1).

\vspace{-2mm}
\paragraph{$\blacktriangleright$ Step 2: 2D-to-4D Scene Transcending Learning.}

In this step, we focus on 2D-to-4D Scene Transcending Learning to enable the transition from 2D to 4D scene. This process is further divided into three substeps:

\textbf{1) RGB-to-Depth Transcending Learning:}
We utilize 200K depth estimation samples from the DIML dataset \cite{DIML-abs-2110-11590} to train the depth estimator. The input comprises 2D RGB images, with corresponding ground-truth depth images providing supervision. The optimization follows Eq. (2).

\textbf{2) RGB Temporal Learning:}
For temporal learning, we employ 288K video data from the AG dataset. The first frame of each video serves as the input 2D RGB image, and the subsequent frames are used as ground-truth supervision for optimizing the RGB Temporal Estimator, guided by Eq. (4).

\textbf{3) Depth Temporal Learning:}
We leverage depth sequences from the PSG4D dataset. Specifically, the first depth image in each sequence is used as the input, and the remaining depth images in the sequence are utilized as ground-truth for optimizing the Depth Temporal Estimator, following Eq. (5).

These three substeps are independent and can be conducted concurrently, ensuring an efficient training process for the 2D-to-4D transcending mechanism.

\vspace{-2mm}
\paragraph{$\blacktriangleright$ Step 3: Pseudo 4D Scene Transfer Initiation Learning.}

In this step, we leverage 3K samples from the PSG4D dataset for learning. 
Specifically, each 4D scene in the training data is firstly decomposed into three components: 3D scenes, video sequences (i.e., RGB sequences), and depth sequences. These components are then used to train the 2D-to-4D Scene Transcending Module, explicitly optimizing the Depth Estimator, RGB Temporal Estimator, and Depth Temporal Estimator.
Secondly, the pseudo-4D scenes generated using the trained 2D-to-4D transcending module serve as input to the 4D-LLM, which predicts the final 4D-PSGs, i.e., SG triplets and mask tubes.
Ground-truth 4D-PSGs are employed as supervision to optimize the 4D-LLM further.
The overall loss function integrates these components, as detailed in Eq. (7), ensuring cohesive optimization across the whole framework.

\vspace{-2mm}
\paragraph{$\blacktriangleright$ Step 4: Large-scale Visual Scene Transfer Learning.}
In this step, we leverage a large-scale dataset consisting of 150K 2D-SG samples, including VG \cite{VG-KrishnaZGJHKCKL17} and PSG \cite{PSG-YangAGZZ022}, to train the 4D-LLM.
The process begins by feeding 2D scenes into the 2D-to-4D Scene Transcending Module, transforming the input into representations suitable for the 4D-LLM. 
The 4D-LLM then interprets these representations and generates the corresponding 2D-PSGs. 
The predicted 2D-PSGs are supervised using the ground-truth 2D-PSGs, ensuring accurate SG generation.

\vspace{-2mm}
\paragraph{$\blacktriangleright$ Step 5: 4D Scene Fine-tuning.}
To ensure optimal model performance, we incorporate an additional training step focused on 4D scene fine-tuning. 
In this step, we repeat the training process outlined in Step 1 using the P4G4D dataset, allowing the model to further refine its understanding of 4D scenes and enhancing its ability to generate accurate 4D-PSGs.

\vspace{-2mm}
\section{System Inference}
\label{app:inference}
\vspace{-1mm}

To improve the quality and address out-of-vocabulary (OOV) issues in 4D-PSG generation, we employ a chained inference mechanism during the inference phase. The inference process is divided into four sequential stages:
\begin{compactitem}
    \item \textbf{Inference stage 1: Object Description and Categorization.} In this stage, the input 4D scene is analyzed to identify all objects present. To handle OOV issues, the LLM first generates detailed descriptions of each object before assigning them specific categories, ensuring a robust recognition, even for unseen or ambiguous objects.
    \item \textbf{Inference stage 2: Semantic Relation Identification.} Based on the identified objects, the LLM determines which object pairs exhibit semantic relationships, establishing the foundation for constructing meaningful scene graphs.
    \item \textbf{Inference stage 3: Precise Relation Description.} To refine the semantic relationships, the LLM generates predicates that offer precise and contextually relevant descriptions of the interactions between object pairs. This step avoids overly general or coarse-grained predicates, ensuring a higher granularity and interpretability.
    \item \textbf{Inference stage 4: Temporal Span Determination.} For object pairs with confirmed semantic relationships, the model further infers the temporal span during which these relationships are valid within the given 4D scene. 
\end{compactitem}
To enhance the model’s reasoning ability and comprehension of complex instructions, we employ in-context learning throughout the chained inference process. 
Below, we detail the prompts used to guide the LLM effectively.
\vspace{-1mm}
\begin{tcolorbox}[breakable, fontupper=\customfont, title=Chained Scene Graph Inference]
\vspace{-2mm}
{\small
\textbf{Input Data}: 4D Scene, the duration \\
\textbf{Instruction}: 
You are a scene expert with professional skills in generating an SG triplets sequence. You follow these four detailed steps to ensure a logical, step-by-step approach to SG generation: \\

\textbf{\color{ired}{Inference stage 1: Object Description and Categorization}}. 
For each object in the scene, do not immediately output its name. Instead, start by describing each object in detail.
Provide a description of each object based on its appearance, shape, structure, and any unique characteristics observed in the scene. 
After giving a detailed description, assign a category to the object that best fits the objects (e.g., ``person'', ``table'', ``chair'', etc.).\\
\textbf{\color{ired}{Expected Output}}: (description, object$_1$), $\cdots$ \\

\textbf{\color{blue}{Inference stage 2: Semantic Relation Identification}}. 
Based on the identified objects, analyze which pairs of objects may have semantic relations. Consider spatial positioning, interactions, and any logical connections that might exist between them.
Identify only pairs that have a meaningful relationship, and briefly explain why these pairs might be related.\\
\textbf{\color{blue}{Expected Output}}: (object$_i$, object$_j$), $\cdots$ \\

\textbf{\color{igreen}{Inference stage 3: Precise Relation Description}}
For each object pair identified in Step 2, describe the exact nature of the relation between the two objects as precisely as possible.
Use clear, concise language to specify the relation type (e.g., "sitting on," "holding," "near," etc.) and provide additional context if necessary to ensure the relation is unambiguous.\\
\textbf{\color{igreen}{Expected Output}}: (object$_i$, relation$_k$ object$_j$), $\cdots$ \\

\textbf{\color{ipurple}{Inference stage 4: Temporal Span Determination}}
For each identified relation, determine its duration or time span. Indicate if the relation is continuous, occurs intermittently, or exists only at a specific moment within the scene.
Use a numerical value for the duration, such as a time interval (e.g., (0.1, 0.7) ) \\
\textbf{\color{ipurple}{Expected Output}}: (object$_i$, relation$_k$ object$_j$, start\_time, end\_time), $\cdots$ \\

\textbf{Final Output Format}:
For each object pair and relation, generate SG triplets in the following format:\\
\textbf{\color{black}{Expected Output}}: (object$_i$, relation$_k$ object$_j$, start\_time, end\_time), $\cdots$
}
\end{tcolorbox}

\begin{tcolorbox}[breakable, fontupper=\customfont, title=In-context Examples]
\vspace{-2mm}
{\small

\textbf{[Example-1]} \\
\textbf{Input Data}: 4D Scene \logo, the duration \\

\textbf{\color{ired}{Inference stage 1: Object Description and Categorization}}. \\
\textbf{Object 1}: \\
\textit{Description}: A person wearing a white T-shirt, gray pants, and black gloves. They are wearing a dark blue helmet and appear to be running on a sandy field.\\
\textit{Category}: Person \\
\textbf{Object 2}:\\
\textit{Description}: A person wearing a green shirt, gray pants, and a green baseball cap. They are holding a baseball glove and appear to be positioned slightly behind Object 1.\\
\textit{Category}: Person\\
\textbf{Object 3:}\\
\textit{Description}: A green outfield area with a mix of trees in the background.\\
\textit{Category}: Field \\
\textbf{Object 4:}\\
\textit{Description}: A sandy dirt area forming the base path of the field.
\textit{Category}: Ground \\

\textbf{\color{blue}{Inference stage 2: Semantic Relation Identification}}. \\
(Person, Person) - The runner (Object 1) is running toward the fielder (Object 2), indicating a potential play interaction. \\
(Person, Ground) - The runner (Object 1) is running along the base path (Object 3).\\
(Person, Ground) - The fielder (Object 2) is positioned on or near the base path (Object 3).\\
(Ground, Field) - The Ground (Object 3) is part of the larger field (Object 4).\\

\textbf{\color{igreen}{Inference stage 3: Precise Relation Description}}\\
(Person 1, running toward, Person 2) \\
(Person 1, running along, Ground) \\
(Person 2, standing near, Ground)\\
(Ground, part of, Field)\\

\textbf{\color{ipurple}{Inference stage 4: Temporal Span Determination}} \\
(Person 1, running toward, Person 2, start time: 0.2, end time: 0.8), (Person 1, running along, Ground, start time: 0.1, end time: 0.9), (Person 2, standing near, Ground, start time: 0.0, end time: 1.0), (Ground, part of, Field, start time: 0.0, end time: 1.0). \\

\textbf{Final Output Format}: \\
(Person 1, running toward, Person 2, 0.2, 0.8)
(Person 1, running along, Ground, 0.1, 0.9)
(Person 2, standing near, Ground, 0.0, 1.0)
(Ground, part of, Field, 0.0, 1.0)
}
\end{tcolorbox}

\section{Dataset Specification}
\label{app:datasets}

\vspace{-1mm}
\paragraph{PSG4D.} This dataset \cite{4d-psg-yang20244d} contains 2 subsets:
\textbf{(a) PSG4D-GTA} selected from the SAIL-VOS 3D \cite{SAIL-VOS-HuWYS21} dataset, containing contains 67 videos with an average length of 84 seconds, amounting to 27,700 RGB-D images, 28.3 billion point clouds, and comprises 35 object categories, and 43 relationship categories;
\textbf{(b) PSG4D-HOI} from HOI4D \cite{HOI4D-LiuLJLWSLFWY22} dataset, including 2,973 videos with an average duration of 20 seconds, equating to 891,000 RGB-D images across 282 indoor scenes. This dataset includes 46 object categories and 15 object-object relationship categories.

\vspace{-3mm}
\paragraph{Visual Genome (VG).} 
We leverage the original VG dataset \cite{VG-KrishnaZGJHKCKL17} for training, which contains the 5,996 types of objects, 1,014 types of predicates, and approximately 108k images.

\vspace{-3mm}
\paragraph{Panoptic Scene Graph (PSG).}
Filtered from COCO \cite{COCO-LinMBHPRDZ14} and VG datasets \cite{VG-KrishnaZGJHKCKL17}, the PSG dataset \cite{PSG-YangAGZZ022} contains 133 object classes, including things, stuff, and 56 relation classes. This dataset has 46k training images and 2k testing images with panoptic segmentation and scene graph annotation. We follow the same data-processing pipelines from \cite{PSG-YangAGZZ022}. 

\vspace{-3mm}
\paragraph{Action Genome (AG).} 
AG \cite{AG-JiK0N20} annotates 234,253 frame scene graphs for sampled frames from around 10K videos, based on the Charades dataset \cite{SigurdssonVWFLG16}. The annotations cover 35 object categories and 25 predicates. The overall predicates consist of three types of predicates: attention, spatial, and contracting.

\vspace{-3mm}
\paragraph{DIML.} DIML \cite{DIML-abs-2110-11590} comprises 2M color images and their corresponding depth maps from a great variety of natural indoor and outdoor scenes. The indoor dataset was constructed using the Microsoft Kinect v2, while the outdoor dataset was built using the stereo cameras (ZED stereo camera and built-in stereo camera). We randomly select the 200K samples for training.

\vspace{-1mm}
\section{Detailed Experimental Implementations}
\label{app:implementations}
\vspace{-1mm}

We employ Imagebind \cite{imagebind-GirdharELSAJM23} as our 4D scene encoder.
Similarly, Imagebind applies the image, video decoder, and depth encoder when performing the 2D-to-4D transfer learning.
The design of the aggregator utilized for fusing RGB and depth features follows in \cite{ChenLWWQLZ20}.  
The projector is implemented as a 2-layer MLP. 
The LLM is instantiated with LLaMA2 \cite{llama2-abs-2307-09288} and fine-tuned using LoRA \cite{lora-HuSWALWWC22}. 
We initialize the mask decoder with SAM2 \cite{sam2-abs-2408-00714} weights. 
The Depth Estimator consists of a $3 \times 3$ convolutional layer and a projector implemented as a $1 \times 1$ convolution layer to predict depth features. 
Both the RGB Temporal Estimator and the Depth Temporal Estimator use 6 transformer layers, with a 512-dimensional embedding dimension, and 8 attention heads. 
The optimizer is AdamW, with an inverse square root learning rate schedule and warm-up steps. 
The training is carried out end-to-end on 8 H100 80GB GPUs with distributed training based on DeepSpeed.
We summarize the training recipes for 4D-LLM in Tab. \ref{tab:training_recipes}.

\begin{table*}[!th]
\centering
\fontsize{8}{10}\selectfont
\setlength{\tabcolsep}{2mm}
\begin{tabular}{lcccccc}
\hline
\multirow{2}{*}{\bf Configuration} & \multirow{2}{*}{\bf Step-1} & \multicolumn{3}{c}{\bf Step-2} & \multirow{2}{*}{\bf Step-3} & \multirow{2}{*}{\bf Step-4 }\\
\cmidrule(r){3-5}
& & \bf Subprocess-a & \bf Subprocess-b & \bf Subprocess-c & & \\

\toprule
Optimizer & AdamW & AdamW & AdamW  & AdamW & AdamW & AdamW  \\
Precision & bfloat16 & bfloat16 & bfloat16 & bfloat16 & bfloat16 & bfloat16 \\
Peak learning rate of LLM & 5e-5 & - & -  & - & 5e-5 & 5e-5 \\
Peak learning rate of Visual Part & 5e-4 & 2e-3 & 5e-3  &  2e-4 & 2e-4 & 5e-4  \\
Weight Decay & 0.05 & 0.1 & 0.1 & 0.1 & 0.05 & 0.05 \\
Learning Rate Scheduler & Cosine & Cosine & Cosine & Cosine & Cosine & Cosine \\
LR Warmup Steps & 500 & 500 & 500 & 500 & 500 & 500  \\
Training Data & PSG4D \cite{4d-psg-yang20244d} & DIML \cite{DIML-abs-2110-11590} &  AG \cite{AG-JiK0N20} & PSG4D \cite{4d-psg-yang20244d} & PSG4D \cite{4d-psg-yang20244d} & VG \cite{VG-KrishnaZGJHKCKL17}, PSG \cite{PSG-YangAGZZ022} \\
\hline
\end{tabular}
\vspace{-2mm}
\captionof{table}{
\label{tab:training_recipes}
Training recipes for 4D-LLM.
}
\vspace{-3mm}
\end{table*}

\vspace{-1mm}
\section{Additional Experiments}
\label{app:experiments}

\vspace{-1mm}
\paragraph{Analyzing 4D-LLM.}
To comprehensively understand the strengths and motivations behind 4D-LLM, it is essential to analyze its design and advantages.

First, the decision to utilize LLMs stems from their inherent richness in knowledge and emergent capabilities. 
LLMs excel in handling diverse textual tasks due to their extensive pretraining, which equips them with a robust internal knowledge base. 
By leveraging this knowledge, 4D-LLM is designed to achieve fine-grained perception across various scenes. 
This aligns with our goal of fully utilizing the internal capabilities of LLMs to address the challenges of 4D-PSG generation, ensuring precise understanding and representation of complex environments.

Another key motivation for using an LLM-based approach is its potential to seamlessly integrate with downstream tasks. 
The actual utility of 4D-PSG lies in its ability to serve broader applications, such as robotic navigation and role-playing simulations. 
LLM-based models are particularly advantageous in these scenarios due to their adaptability and capacity for rapid transfer to new tasks. 
This flexibility ensures that 4D-LLM can maximize its impact beyond PSG generation, making it an ideal candidate for practical applications in dynamic, multimodal environments.

Experimentally, we have demonstrated that the LLM-based 4D-PSG generation method achieves substantial performance improvements over baseline models. 
These gains validate the efficacy of incorporating LLMs into 4D scene understanding tasks. 
Furthermore, by integrating our innovative 2D-to-4D visual scene transfer learning approach, we observed additional performance enhancements. 
This highlights the synergy between LLM-based architectures and our novel methods, underscoring the effectiveness of 4D-LLM in advancing 4D scene graph generation and its broader applicability.

\begin{table}[t]
\fontsize{8}{10}\selectfont
\setlength{\tabcolsep}{1.9mm}
\centering
\begin{tabular}{lcccc}
    \toprule
    \multirow{1}{*}{\bf Method}  & \bf R@50 & \bf R@100  & \bf mR@50 & \bf mR@100 \\
    \midrule
    \multicolumn{5}{l}{$\bullet$ Supervised learning} \\
    Motifs \cite{Motifs-ZellersYTC18} & \bf 28.9 & \bf 33.1 & 6.4 & 7.7 \\
    Motifs + CFA \cite{CFA-LiCXYW023} & - & - & \bf 11.6 & \bf 13.2 \\
    VCTree \cite{VCTree-TangZWLL19} & 28.3 & 31.9 & 6.5 & 7.4 \\
    VETO \cite{VETO-SudhakaranDK023} & 26.1 & 29.0  & 7.0 & 8.1 \\
    \cdashline{1-5}
    \multicolumn{5}{l}{$\bullet$ Zero-shot setting} \\
    4D-LLM (ours)  & 28.0  & 32.3  & 10.9  &  13.1  \\
    \bottomrule
\end{tabular}
\vspace{-2mm}
\caption{Zero-shot 2D image SG generation performance on GQA \cite{GQA-HudsonM19} dataset.}
\label{tab:GQA}
\vspace{-2mm}
\end{table}

\begin{table}[t]
\fontsize{8}{11}\selectfont
\setlength{\tabcolsep}{0.8mm}
\centering
\begin{tabular}{lcccc}
    \toprule
    
    \multirow{2}{*}{\bf Method} & \multicolumn{2}{c}{\bf PSG4D-GTA} & \multicolumn{2}{c}{\bf PSG4D-HOI} \\
    \cmidrule(lr){2-3} \cmidrule(lr){4-5}
     & \bf R/mR@20 & \bf R/mR@50 & \bf R/mR@20 & \bf R/mR@50 \\
    \midrule
    PSG4DFormer\textsuperscript{one-stage} & \multirow{2}{*}{6.47 / 3.56} & \multirow{2}{*}{6.85 / 3.01} & \multirow{2}{*}{5.42 / 3.78} & \multirow{2}{*}{5.86 / 3.45} \\
     \quad + $V^{2\rightarrow4}$-{VST} & &  &  & \\
     \cdashline{1-5}
    4D-LLM w/o PSG4D & 8.45 / 5.03 & 9.09 / 5.32 & 6.45 / 4.01 & 8.45 / 6.79 \\
    \bottomrule
\end{tabular}
\vspace{-3mm}
\caption{Zero-shot analysis: The PSG4D dataset is excluded from training, with only 4D data used for 2D-to-4D scene transcending.}
\label{tab:zero-shot}
\vspace{-2mm}
\end{table}

\begin{table}[t]
\fontsize{8}{10}\selectfont
\setlength{\tabcolsep}{2.5mm}
\centering
\begin{tabular}{lcccc}
    \toprule
     \multirow{2}{*}{\bf Method} & \multicolumn{2}{c}{\bf With Constraint}  & \multicolumn{2}{c}{\bf No Constraint} \\
 \cmidrule(r){2-3}\cmidrule(r){4-5}
 & \bf R@20 &\bf  R@50  & \bf R@20  & \bf R@50 \\
    \midrule
    \multicolumn{5}{l}{$\bullet$ Supervised learning} \\
    VCTree~\cite{VCTree-TangZWLL19} & 32.6  & 34.7 & 35.3 & 46.8  \\
    GPS-Net~\cite{GPS-Net-LinDZT20} & 33.1  & 35.1 &   35.7 & 47.3  \\
    STTran~\cite{STTran-CongLARY21} &     34.1 & \bf 37.0 & 36.2  & \bf 48.8  \\
    \cdashline{1-5}
    \multicolumn{5}{l}{$\bullet$ Zero-shot setting} \\
    4D-LLM (ours)  & \bf 34.8   & 36.1 & \bf 40.9  & 48.3   \\
    \bottomrule
\end{tabular}
\vspace{-2mm}
\caption{Zero-shot 2D video SG generation performance on AG \cite{AG-JiK0N20} dataset.}
\label{tab:AG}
\vspace{-2mm}
\end{table}

\vspace{-2mm}
\paragraph{The Generalization Capability.}

We further evaluate the generalization capability of the proposed 4D-LLM by testing its performance on SG generation tasks for 2D images and videos. 
Specifically, we conduct zero-shot inference using 4D-LLM on the GQA \cite{GQA-HudsonM19} and AG \cite{AG-JiK0N20} datasets, with results presented in Tab. \ref{tab:GQA} and \ref{tab:AG}.
Compared to fully supervised baselines, our model achieves comparable performance in 2D image and video SG generation tasks. 
This demonstrates the strong generalization ability of 4D-LLM, which not only excels in 4D-PSG generation but also performs effectively across other modalities. 
These findings further highlight the model's versatility in handling various SG generation tasks, making it applicable to various scenarios, such as robot navigation and other downstream tasks.

\vspace{-2mm}
\paragraph{Detailed Example of Chained Inference Results.}

Here, we show two detailed examples of chained inference results.

\begin{tcolorbox}[breakable, fontupper=\customfont, title=Detailed results of the chained inference process for the example shown in Fig. 5]
\vspace{-2mm}
{\small

\textbf{Input Data}: 4D Scene \logo, the duration \\

\textbf{\color{ired}{Inference stage 1: Object Description and Categorization}}. \\
\textbf{Object 1}: \\
\textit{Description}: A pair of hand tools with metal jaws and a handle wrapped in a white material. It has a pivot point that allows it to grip or manipulate objects. \\
\textit{Category}: Pliers\\

\textbf{Object 2}: \\
\textit{Description}: A rectangular, flat surface with a smooth, glossy finish, decorated with an image of two reindeer and text. Likely a piece of furniture used to hold items.\\
\textit{Category}: Table\\

\textbf{Object 3}:\\
\textit{Description}: A hand with a visible wrist, reaching toward the pliers, likely indicating an interaction with them.\\
\textit{Category}: Hand\\

\textbf{Object 4}:\\
\textit{Description}: A soft cushion with a yellow fabric cover, lying on a blue quilted surface with floral patterns.\\
\textit{Category}: Pillow\\

\textbf{Object 5}:\\
\textit{Description}: A large, cushioned seating area with a fabric cover in a neutral color, placed near the table and likely supporting the pillow.
\textit{Category}: Sofa\\

\textbf{\color{blue}{Inference stage 2: Semantic Relation Identification}}. \\
(Hand, Pliers) - The hand is reaching toward or interacting with the pliers, suggesting a possible action or manipulation. \\
(Pliers, Table) - The pliers are resting on the table surface, indicating a positional relationship. \\
(Pillow, Table) - The pillow is lying on the table or adjacent furniture, indicating a nearby or resting position. \\
(Sofa, Table) - The sofa is positioned near the table, indicating a spatial relationship.\\

\textbf{\color{igreen}{Inference stage 3: Precise Relation Description}}\\
(Hand, reaching toward, Pliers)\\
(Hand, picking up, Pliers)\\
(Hand, holding, Pliers)\\
(Pliers, resting on, Table) \\
(Pillow, lying near, Table) \\
(Sofa, positioned near, Table)\\

\textbf{\color{ipurple}{Inference stage 4: Temporal Span Determination}} \\
(Hand, reaching toward, Pliers, start time: 0.0, end time: 0.2)(Hand, picking up, Pliers, start time: 0.2, end time: 0.55)(Hand, holding, Pliers, start time: 0.5, end time: 1.0) (Pliers, resting on, Table, start time: 0.0, end time: 0.4) (Pillow, lying near, Table, start time: 0.0, end time: 1.0) (Sofa, positioned near, Table, start time: 0.0, end time: 1.0)\\

\textbf{Final Output Format}: \\
(Hand, reaching toward, Pliers, 0.0, 0.2) (Hand, picking up, Pliers, 0.2, 0.55) (Hand, holding, Pliers, 0.5, 1.0) (Pliers, resting on, Table, 0.0, 0.4) (Pillow, lying near, Table, 0.0, 1.0) (Sofa, positioned near, Table, 0.0, 1.0)
}
\end{tcolorbox}

\begin{figure*}[!th]
    \centering
    \includegraphics[width=0.80\linewidth]{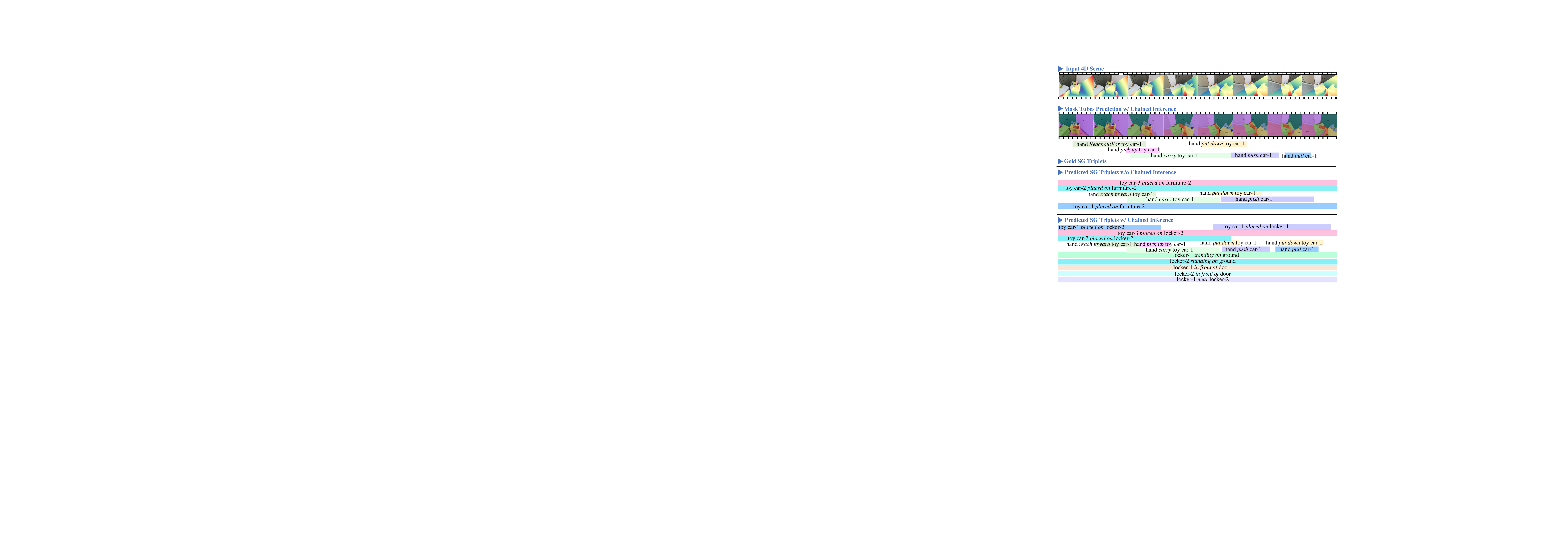}
    \caption{A case illustrating the prediction of 4D-LLM on 4D-PSG}
    \label{fig:case2}
\end{figure*}

\begin{tcolorbox}[breakable, fontupper=\customfont, title=Detailed results of the chained inference process for the example shown in Fig. 6]
\vspace{-2mm}
{\small

\textbf{Input Data}: 4D Scene \logo, the duration \\

\textbf{\color{ired}{Inference stage 1: Object Description and Categorization}}. \\
\textbf{Object 1}: \\
\textit{Description}: A person with short black hair, wearing a blue shirt, yellow pants, and gloves, standing on or near a railroad track, facing another person. \\
\textit{Category}: Person\\

\textbf{Object 2}: \\
\textit{Description}: A second person with short hair, wearing a dark jacket, bending forward near the railroad track, partially obscured in the foreground.\\
\textit{Category}: Person\\

\textbf{Object 3}:\\
\textit{Description}: A set of metal railroad tracks, extending through the scene, with parallel rails and wooden ties.\\
\textit{Category}: Railroad Track\\

\textbf{Object 4}:\\
\textit{Description}: A low concrete barrier running alongside the railroad track, separating the path from the track area.
\textit{Category}: Road Barrier

\textbf{Object 5}: \\
\textit{Description}: An industrial structure in the background, with metal and concrete components, possibly part of a factory or warehouse. \\
\textit{Category}: Industrial Building \\

\textbf{Object 6}: \\
\textit{Description}: There are irregularly shaped stones covering the ground near the railway tracks. \\
\textit{Category}: gravel \\

\textbf{\color{blue}{Inference stage 2: Semantic Relation Identification}}. \\
(Person 1, Person 2) - Person 1 and Person 2 are positioned close to each other, suggesting a possible interaction or confrontation.\\
(Person 1, Railroad Track) - Person 1 is standing near or on the railroad track, indicating a spatial relationship. \\
(Person 2, Railroad Track) - Person 2 is positioned close to the railroad track, indicating proximity. \\
(Person 1, Road Barrier) - The person is positioned near the road barrier, suggesting a spatial relationship. \\
(Person 1, gravel) - The person is positioned on the gravel, suggesting a spatial relationship. \\
(Railroad Track, Industrial Building) - The railroad track leads towards or is located near the industrial building, indicating spatial context. \\

\textbf{\color{igreen}{Inference stage 3: Precise Relation Description}}\\
(Person 1, in front of, Person 2) \\
(Person 1, talking to, Person 2) \\
(Person 1, walking on, Gravel) \\
(Person 1, behind, Road Barrier)\\
(Person 1, walking alongside, Railroad Track)\\
(Person 2, walking alongside, Railroad Track)\\
(Railroad Track, near, Industrial Building)\\

\textbf{\color{ipurple}{Inference stage 4: Temporal Span Determination}} \\
(Person 1, in front of, Person 2, start time: 0.8, end time: 1.0) \\
(Person 1, talking to, Person 2, start time: 0.8, end time: 1.0) \\
(Person 1, walking on, Gravel, start time: 0, end time: 0.2) \\
(Person 1, behind, Road Barrier, start time: 0, end time: 0.2)\\
(Person 1, walking alongside, Railroad Track,  start time: 0.2, end time: 0.8)\\
(Person 2, walking alongside, Railroad Track,  start time: 0.8, end time: 1.0)\\
(Railroad Track, near, Industrial Building, start time: 0.1, end time: 0.6)\\

\textbf{Final Output Format}: \\
(Person 1, in front of, Person 2, 0.8, 1.0),(Person 1, talking to, Person 2, 0.8, 1.0), (Person 1, walking on, Gravel, 0, 0.2), (Person 1, behind, Road Barrier, 0, 0.2), (Person 1, walking alongside, Railroad Track,  0.2, 0.8), (Person 2, walking alongside, Railroad Track,  0.8, 1.0), (Railroad Track, near, Industrial Building, 0.1, 0.6)
}
\end{tcolorbox}

\vspace{-2mm}
\paragraph{More Visualizations.}

Fig. \ref{fig:case2} provides an additional example of predictions generated by 4D-LLM, illustrating its ability to identify both semantic actions and fine-grained spatial relationships within the scene. For instance, the model successfully captures interactions such as ``toy-car \textit{placed on} furniture-2'', ``locker-1 \textit{standing on\textbf{}} ground''. 
NNotably, with the integration of the chained inference mechanism, 4D-LLM has significantly enhanced its ability to distinguish finer object details, providing more robust performance in differentiating ``furniture'' from ``locker''.
Additionally, the chained inference mechanism enhances the model's accuracy in recognizing key semantic relationships over extended timeframes, further improving its performance in scenarios involving long-duration activities.
These results emphasize the robust capability of 4D-LLM for a detailed and precise understanding of complex 4D environments, validating its effectiveness in generating high-quality 4D-PSGs.


